\documentclass[journal]{IEEEtran}
\ifCLASSINFOpdf
\else
\fi
\usepackage{amssymb}
\usepackage{amsmath}
\usepackage{algorithm}
\usepackage{algorithmic}
\usepackage{hyperref}
\usepackage{tikz}
\def\checkmark{\tikz\fill[scale=0.4](0,.35) -- (.25,0) -- (1,.7) -- (.25,.15) -- cycle;} 
\usepackage{graphicx}
\usepackage{booktabs}
\usepackage{xspace}

\usepackage{multirow}
\usepackage{makecell}

\usepackage{ mathrsfs }
\usepackage{graphicx, adjustbox}
\usepackage{array}
\usepackage{setspace}
\usepackage{enumitem}
\usepackage{breqn}
\usepackage{bm}
\usepackage{ dsfont }
\usepackage{ upgreek }
\usepackage{multirow}
\usepackage{algorithm}
\usepackage{algorithmic}
\usepackage{caption}
\usepackage{pgfplots}
\usepackage{tkz-euclide}
\usepackage{subcaption}
\usepackage{tikz}{\tiny }
\usepackage{balance}
\usepackage{ tipa }

\usepackage{fancyhdr}
\usepackage{ upgreek }
\usepackage{soul}
\usepackage{breqn}
\usepackage[english]{babel}
\usepackage{braket} 

\addto\captionsenglish{}
\allowdisplaybreaks

\usepackage{amsthm}

\usepackage{fancyhdr}
\usepackage{ upgreek }
\usepackage{soul}
\makeatletter
\newcommand{\vast}{\bBigg@{4}}

\newcommand{\Vast}{\bBigg@{5}}
\begin{document}

\title{Towards Personalized Quantum Federated Learning for Anomaly Detection}

\author{Ratun Rahman, Sina Shaham, 
        and Dinh C. Nguyen,~\IEEEmembership{Member,~IEEE}

\thanks{Ratun Rahman and Dinh C Nguyen are with the Department of Electrical and Computer Engineering, University of Alabama in Huntsville, Huntsville, AL 35899, USA, emails: rr0110@uah.edu, dinh.nguyen@uah.edu} 
\thanks{Sina Shaham is at Meta, San Francisco Bay Area, CA, USA, email: sshaham@usc.edu} }



\maketitle

\begin{abstract}
Anomaly detection has a significant impact on applications such as video surveillance, medical diagnostics, and industrial monitoring, where anomalies frequently depend on context and anomaly-labeled data are limited.  Quantum machine learning (QML) offers powerful tools for effectively processing high-dimensional data, but centralized QML systems face considerable challenges, including data privacy concerns and the need for massive quantum resources at a single node.  Quantum federated learning (QFL) overcomes these concerns by distributing model training among several quantum clients, consequently eliminating the requirement for centralized quantum storage and processing.
However, in real-life quantum networks, clients frequently differ in terms of hardware capabilities, circuit designs, noise levels, and how classical data is encoded or preprocessed into quantum states. These differences create inherent heterogeneity across clients—not just in their data distributions, but also in their quantum processing behaviors. As a result, training a single global model becomes ineffective, especially when clients handle imbalanced or non-identically distributed (non-IID) data.
To address this, we propose a new framework called personalized quantum federated learning (PQFL) for anomaly detection. PQFL enhances local model training at quantum clients using parameterized quantum circuits and classical optimizers, while introducing a quantum-centric personalization strategy that adapts each client’s model to its own hardware characteristics and data representation. This balances local customization with global coordination.
Extensive experiments show that PQFL significantly improves anomaly detection accuracy under diverse and realistic conditions. Compared to state-of-the-art methods, PQFL reduces false errors by up to 23\%, and achieves gains of 24.2\% in AUROC and 20.5\% in AUPR, highlighting its effectiveness and scalability in practical quantum federated settings.

\end{abstract}

\maketitle



\section{Introduction}\label{sec:introduction}
\IEEEPARstart{A}{nomaly} detection is crucial in high-risk applications such as medical diagnostics, industrial safety, cyberattack defense, and video surveillance, where detecting rare and context-specific deviations can avoid costly or fatal consequences \cite{wang2024regression,bhuyan2013network}. {\color{black}In medical diagnostics, for example, failing to identify an anomaly could result in overlooking a vital condition like cancer or heart disease, leading to delayed treatment and potentially life-threatening effects. In industrial environments, unnoticed anomalies in machinery or processes can cause equipment failure, production downtime, or even catastrophic accidents. Similarly, in cyberattack protection, abnormalities frequently signify security breaches or malicious activity that, if not immediately addressed, might compromise important data and systems. \cite{alanazi2023anomaly, cui2019machine}. Video surveillance systems use anomaly detection to identify abnormal behaviors or events that could suggest criminal activity or safety issues \cite{zhou2019anomalynet, patrikar2022anomaly}. Thus, the ability to reliably and efficiently detect anomalies is not simply a technical challenge, but a subject with significant societal and economic importance. \cite{zhou2021feature}.}

However, anomaly identification remains a fundamentally challenging task due to \textit{two major factors} \cite{yan2023fgdae}: (i) anomalies are frequently described in terms of local context rather than global trends \cite{li2023deep, bahri2022automl}, and (ii) labeled anomalous samples are limited or largely unavailable in practical situations \cite{kim2022graph,pang2021deep}. {\color{black}The first factor complicates detection because anomalies often manifest as subtle, context-dependent deviations that are difficult to distinguish from normal variations or noise \cite{shen2025early}. For example, in a manufacturing process, a slight deviation in temperature might be normal in one context but indicative of a fault in another \cite{li2025application,zhang2023data}. The second factor exacerbates the problem, as machine learning models typically require a substantial amount of labeled data to learn effectively \cite{menghani2023efficient, alzubaidi2023survey}. In anomaly detection, however, anomalous events are rare by definition, making it challenging to collect sufficient labeled examples for training \cite{bergmann2021mvtec,doshi2022rethinking}. This scarcity of data forces models to rely heavily on assumptions about what constitutes normal behavior, increasing the risk of false positives or missed anomalies.}
These issues are exacerbated in distributed scenarios, where data privacy concerns and transmission overheads make centralized training impractical \cite{dong2024fadngs}. Recent improvements in quantum machine learning (QML) \cite{wang2022data,kukliansky2024network} provide an exciting prospect for high-dimensional data analysis. {\color{black}In centralized QML frameworks, data from numerous sources must be aggregated at a central server, which raises serious privacy concerns \cite{cerezo2022challenges, caro2022generalization}, particularly when dealing with sensitive information \cite{rajawat2023quantum} such as medical records or private industrial data.  Furthermore, the quantum gear required to handle large-scale datasets centrally is frequently prohibitively expensive and may be out of reach for many businesses \cite{watkins2023quantum}.  These constraints restrict the effective deployment of QML in real-world applications with data distributed across numerous locations or devices \cite{ren2025toward}.}

To address the limitations of centralized quantum learning, quantum federated learning (QFL) has emerged as an effective approach for distributing quantum model training over numerous clients, each client having local quantum processors \cite{wang2024quantum}.  QFL keeps sensitive data on-device and only shares quantum model updates with a central server, protecting privacy while decreasing the requirement for resource-intensive centralized quantum resources \cite{dong2024fadngs,pokharel2025quantum}.  However, current QFL techniques use the assumption that all clients run under similar hardware circumstances and use identical quantum encoding algorithms, which rarely applies in practice. Clients in realistic quantum networks may have varying circuit layouts, noise profiles, encoding techniques, and data preparation processes \cite{pokharel2025quantum, innan2024fedqnn,chehimi2023foundations}.  These discrepancies result in inherent heterogeneity in both data representation and local quantum model evolution, significantly limiting the efficiency of naïve global model aggregation \cite{gurung2025performance, gurung2024personalized}. 
As a result, standard QFL approaches frequently underperform in non-IID situations, particularly in tasks such as anomaly detection, where small fluctuations can carry significant weights.

\textbf{Main Contributions:} To address the obstacles that arise for diverse quantum clients, we introduce the personalized quantum federated learning (PQFL) framework, which is particularly intended for anomaly detection in real quantum networks.  Unlike traditional QFL techniques, which seek to fit a single global model across all clients, PQFL includes a quantum-centric personalization layer that enables each client to train and maintain a model that is suited to its specific encoding scheme, circuit depth, and data distribution. \textcolor{black}{As a result, this study provides conceptual and methodological contributions:  Instead of claiming a precise quantum computing advantage, PQFL shows how tailored federated techniques may be applied to noisy quantum situations.} The main contributions of this paper are summarized as follows. 

\begin{itemize}
    \item We propose a novel \textit{PQFL} approach in anomaly detection that can effectively handle heterogeneous clients due to \textcolor{black}{heterogeneous quantum encoding} in the QFL environment, ensuring stable performance of the quantum global model. 
    \item \textcolor{black}{We create a \textit{QFL} framework that \textit{synergistically links a regularization-based personalization rule} with quantum feature encoding, allowing for resilience under non-IID data distributions, quicker convergence, and high-speed anomaly discrimination.  This methodological approach emphasizes the combined characteristics of quantum representation learning and federated optimization, resulting in both privacy protection and effective use of limited quantum resources.}
    \item We evaluate our model's performance with other state-of-the-art methods through extensive simulation results. The simulation results indicate that the proposed \textit{PQFL} method improves by up to 23.4\%, 19\%, 24.2\%, and 20.5\% in terms of false error (FE), missing error (ME), area under the receiver operating characteristic curve (AUROC) and area under the precision-recall curve (AUPR), respectively.
\end{itemize}
\textbf{Paper Organization:} The rest of this paper is organized as follows. The related literature is reviewed in Section II, and Section III provides our proposed \textit{PQFL} method. Our simulation results are presented in Section IV. Finally, \textcolor{black}{Section V presents the limitations} and Section VI concludes the paper. 
The key notation of the paper is summarized in Table \ref{tab: notation}.

\begin{table}
\centering
\footnotesize
\caption{List of key notations.}
\begin{tabular}{|p{0.8cm}|p{2.6cm}||p{0.8cm}|p{2.6cm}|}
\hline
\textbf{Notation} & \textbf{Definition} & \textbf{Notation} & \textbf{Definition} \\ 
\hline
$N$ & Number of clients & $K$ & Number of global rounds \\ 
$T$ & Number of local rounds & $l$ & Number of quantum layers\\ 
$D_q$ & Number of qubits & $I$ & Identity matrix \\
$D_n^k$ & clients’s Data size & $U$ & Unitary operation \\
$\boldsymbol{w}_{n,k}^t$ & Model parameter & $O$ & Observed gate \\
$|\psi\rangle$ & Quantum state & $\mathcal{H}$ & Hilbert space \\
$M$ & Number of measurements  & $P_m$ & Projectors on measurement $m$\\
$\lambda$ & Eigenspace  & $\operatorname{Tr}$ & Trace of matrix\\
$\Pi_y$ & Projection operator & $\hat{g}_t$ & Gradient estimate \\

\hline
\end{tabular}
\label{tab: notation}
\end{table}

\section{Related Works}
We present a summary of related studies from a variety of perspectives relevant to our PQFL development for anomaly detection.

\subsection{Existing Anomaly Detection Methods} 
Anomaly detection is a critical task in various domains, with three primary approaches identified in the literature \cite{dong2024fadngs}: (1) \textit{Reconstruction-based methods} (RE), which focus on reconstructing input data to identify anomalies \cite{yan2023fgdae,wang2024regression}; (2) \textit{Classification-based approaches} (CF), which employ supervised learning to distinguish between normal and abnormal data \cite{cao2024adaclip}; and (3) \textit{Constructive learning} (CL) methods, which incrementally build models to capture normal behavior patterns, identifying deviations as anomalies \cite{he2024mambaad, zhang2024realnet}. {\color{black}These methodologies have been extensively investigated across applications such as cybersecurity, medical diagnostics, and industrial monitoring. Each offers unique capabilities but also faces specific obstacles depending on data qualities and operational restrictions.}

A widely adopted RE method is the auto-encoder-based FGDAE proposed in \cite{yan2023fgdae}, which reconstructs input data to learn the normal distribution. However, \textit{this approach requires substantial amounts of data, limiting its applicability in data-scarce scenarios} \cite{yan2023fgdae}. Similarly, RegAD, a recent federated learning (FL) algorithm for anomaly detection, employs an RE approach but \textit{suffers from privacy concerns and increased communication latency due to data sharing requirements} \cite{wang2024regression}. {\color{black}These restrictions emphasize the necessity for privacy-preserving technologies that can work efficiently in distributed systems, especially when centralized data collecting is impracticable \cite{uddin2025false, chen2020joint, pillutla2022robust}.}

CF approaches, often based on support vector machines (SVMs), are simpler to implement and interpret. For instance, \cite{lane1997application} applied a single-class SVM for anomaly detection, but \textit{it was found unsuitable for high-dimensional data due to scalability issues}. To address this, \cite{cao2024adaclip} proposed a deep learning-based single-class SVM, yet \textit{its performance remained suboptimal compared to other methods}. Advanced CF methods like Generalad \cite{strater2024generalad} and Real-iad \cite{wang2024real} leverage deep learning but \textit{require data sharing, raising privacy concerns in distributed settings}. Similarly, MOCCA, a multilayer one-class classification approach \cite{massoli2021mocca}, optimizes feature spaces for anomaly detection but \textit{struggles with non-IID data, limiting its effectiveness in federated scenarios}. {\color{black}These problems highlight the difficulty of implementing standard anomaly detection methods in distributed environments where data heterogeneity and privacy constraints are important impediments \cite{rahman2025electrical, li2020federated}.}

CL methods generally outperform RE and CF approaches by learning discriminative representations, particularly for near-distribution anomalies \cite{hendrycks2019using}. The seminal work in \cite{golan2018deep} introduced a CF-based method using geometric transformations, but \textit{it struggles with limited data in distributed learning settings}. SimCLR \cite{chen2020simple} employs image augmentations to create positive and negative samples, enhancing model robustness. Similarly, Mambaad \cite{he2024mambaad} uses distributionally shifted augmentations to improve discrimination between normal and anomalous samples. Building on this, Realnet \cite{zhang2024realnet} proposed a semi-supervised CL strategy, achieving comparable performance. Inspired by \cite{he2024mambaad, zhang2024realnet}, our work proposes a novel CL-based method that maps normal and anomalous samples into a discriminative feature space, enhancing anomaly detection in federated settings. {\color{black}By utilizing CL's ability to handle complicated data distributions, our technique tackles the limitations of previous methods, particularly in the setting of distributed and heterogeneous data environments \cite{imrey2020limitations}.

However, these algorithms frequently struggle with high-dimensional or unstructured data, such as photos, sensor logs, or videos, due to their reliance on handmade features and the necessity for large labeled data \cite{chalapathy2019deep}.  As data complexity grows, traditional approaches face difficulties in accuracy and processing efficiency.  This drives the development of quantum machine learning approaches, which provide potential advantages in rapidly analyzing high-dimensional data and overcoming computational constraints.}

\subsection{Federated Learning for Anomaly Detection}
{\color{black}Federated learning (FL) has emerged as a promising approach for anomaly detection in distributed environments, enabling collaborative model training without sharing sensitive raw data \cite{konevcny2016federated, 11079607}}. Early work by \cite{zhao2019multi} proposed a classification-based FL method for network anomaly detection, addressing data scarcity by simultaneously training three related tasks to improve model generalization. However, \textit{this approach is unsuitable for unsupervised learning, which is often required in real-world applications where labeled anomalies are scarce}. The works in \cite{qian2025reconstruction} and \cite{yang2025improved} explored reconstruction-based FL techniques for unsupervised anomaly detection, setting a threshold for the global model’s reconstruction error to identify anomalies. These methods operate without requiring labeled data, making them suitable for privacy-sensitive scenarios. Similarly, the state-of-the-art method FADngs \cite{dong2024fadngs} introduced an unsupervised FL-based anomaly detection approach leveraging the concept of false and missing data to enhance detection accuracy. However, reconstruction-based approaches generally suffer with high communication overheads due to frequent model changes and may fail to handle non-IID data well, as local data distributions differ between clients. \textcolor{black}{Furthermore, their dependence on consistent model designs restricts their flexibility to various client settings, such as those with different processing capacities or data characteristics \cite{su2025apfed, cho2021personalized,rahman2025electrical}.}

{\color{black}Recently, research attempted to overcome these limitations by adding sophisticated anomaly detection algorithms into FL. For example, \cite{gurung2025performance} suggested a federated autoencoder system that adapts to non-IID input by providing local model personalization, although it still faces large communication costs. Similarly, \cite{innan2024fedqnn} created a hybrid FL technique combining reconstruction and contrastive learning to enhance anomaly detection in dispersed sensor networks. This approach is resilient against data heterogeneity but requires sophisticated hyperparameter tuning. These developments highlight the growing interest in privacy-preserving anomaly detection, but they fall short of addressing the unique challenges of quantum computing environments, where hardware heterogeneity and quantum-specific constraints (e.g., circuit depth, noise profiles) further complicate federated training \cite{rahman2025sporadic}. Given the practical importance of this problem, our study focuses on unsupervised anomaly detection with a low risk of compromising privacy, leveraging the PQFL framework to address both data and hardware heterogeneity in quantum networks.}

\subsection{Quantum Machine Learning}
Quantum machine learning (QML) integrates quantum computing with machine learning to enhance data processing and model training efficiency, particularly for complex computational tasks \cite{senokosov2024quantum}. QML leverages quantum algorithms to achieve potential speedups over classical methods, especially in large-scale data processing and optimization \cite{biamonte2017quantum}. For instance, \cite{mangini2022quantum,kukliansky2024network} developed a Quantum Neural Network (QNN) for industrial quality control, combining a classifier with an auto-encoder for feature extraction and dimensionality reduction. Similarly, \cite{wang2022data} utilized QNNs to recover missing or corrupted data, emphasizing the importance of accurate quantum state preparation. In \cite{zhang2022quantum}, quantum autoencoders were used to model Hamiltonian dynamics, reducing computational resource requirements by exploiting quantum system properties. Additionally, \cite{dang2018image} demonstrated improved image classification efficiency using a quantum k-nearest neighbor approach, leveraging quantum superposition and minimum search algorithms to achieve computational speedups while maintaining accuracy comparable to classical methods. These developments demonstrate QML's potential for changing computationally intensive tasks by enhancing precision and decreasing resource consumption \cite{schuld2015introduction}.  \textcolor{black}{However, QML is limited by the necessity for centralized data processing, raising privacy problems and requiring large quantum resources \cite{havlivcek2019supervised}, including anomaly detection~\cite{zhu2023anomaly}}. These constraints need decentralized techniques like quantum federated learning to enable realistic deployment in privacy-sensitive and resource-constrained contexts\cite{dunjko2018machine}.

\begin{table*}[ht]
\color{black}
\centering
\caption{Comparison of Federated Learning Approaches for Anomaly Detection}
\label{tab:fl_comparison}
\setlength{\tabcolsep}{4pt}
\renewcommand{\arraystretch}{1.1}
\begin{tabular}{@{\hskip 3pt}l@{\hskip 6pt}c@{\hskip 3pt}c@{\hskip 3pt}c@{\hskip 3pt}c@{\hskip 3pt}c@{\hskip 3pt}c@{\hskip 3pt}c@{\hskip 3pt}}
\toprule
\textbf{Method} &
\shortstack{\textbf{Handles}\\\textbf{Heterogeneity}} &
\shortstack{\textbf{Calculation}\\\textbf{Speed}} &
\shortstack{\textbf{Data}\\\textbf{Privacy}} &
\shortstack{\textbf{Scalability}} &
\shortstack{\textbf{Comm.}\\\textbf{Efficiency}} &
\shortstack{\textbf{Quantum}\\\textbf{Compatible}} &
\shortstack{\textbf{Unsupervised}\\\textbf{Learning}} \\
\midrule
Zhao et al. \cite{zhao2019multi}            & -          & slow     & \checkmark & \checkmark & -          & -          & - \\
quan et al. \cite{qian2025reconstruction}        & -          & slow     & \checkmark & \checkmark & -          & -          & \checkmark \\
Yang et al. \cite{yang2025improved} & -        & slow     & \checkmark & \checkmark & -          & -          & \checkmark \\
FADngs \cite{dong2024fadngs}                & -          & medium   & \checkmark & \checkmark & \checkmark & -          & \checkmark \\
FGDAE \cite{yan2023fgdae}             & -          & medium   & -          & -          & -          & -          & \checkmark \\
RegAD \cite{wang2024regression}             & -          & fast     & -          & -          & -          & -          & \checkmark \\
Adaclip \cite{cao2024adaclip}              & -          & fast     & -          & -          & -          & -          & \checkmark \\
Generalad \cite{strater2024generalad}          & -          & fast     & -          & -          & -          & -          & \checkmark \\
Real-iad \cite{wang2024real}                 & -          & medium   & -          & -          & -          & -          & \checkmark \\
Mambaad \cite{he2024mambaad}                      & -          & medium   & -          & -          & -          & -          & \checkmark \\
Realnet \cite{zhang2024realnet}                     & -          & medium   & -          & -          & -          & -          & \checkmark \\
PFL \cite{su2025apfed, cho2021personalized}                     & \checkmark         & medium   & \checkmark          & \checkmark          & -          & -          & \checkmark \\
QFL                                         & - & fastest  & \checkmark & \checkmark & \checkmark & \checkmark & \checkmark \\
\textbf{PQFL (Ours)}                        & \checkmark & fastest  & \checkmark & \checkmark & \checkmark & \checkmark & \checkmark \\
\bottomrule
\end{tabular}
\end{table*}

\subsection{Quantum Federated Learning}
Quantum Federated Learning (QFL) extends the principles of classical federated learning to the quantum domain, enabling multiple quantum clients with local quantum processors to collaboratively train a global quantum model without sharing raw data \cite{nguyen2025quantum}. This decentralized approach preserves privacy, reduces communication costs, and eliminates the need for large-scale quantum processing at a single location \cite{li2021quantum}. Early work by \cite{chen2021federated} introduced QFL using blind quantum computing, where a client with limited quantum capabilities delegates computations to a server while maintaining data confidentiality. However, the reliance on complex brickwork state preparation posed practical challenges \cite{li2021quantum}. Building on this, \cite{yang2022brickwork} proposed a hybrid QFL system, though its quantum advantages were limited by reliance on classical deep learning for large-scale training. A more robust QFL framework was presented in \cite{rahman2025simqfl}, improving scalability and efficiency.

\begin{figure}
    \centering
    \includegraphics[width=0.99\linewidth]{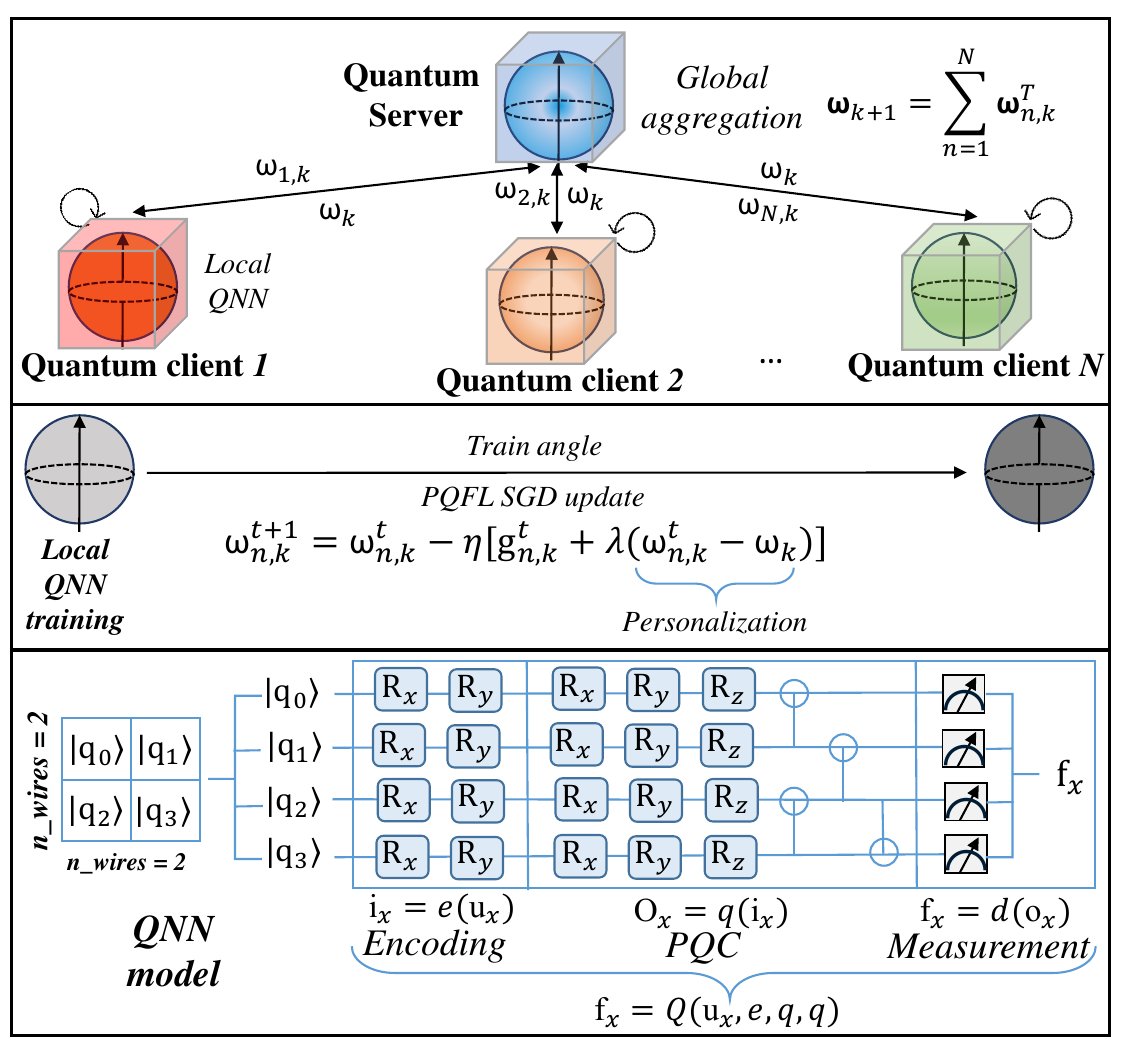}
    \caption{Overview of the \textit{PQFL} architecture with $N$ quantum clients and a classical server within the QFL framework.  Each client encodes conventional input into quantum states and is trained using local QML models with \textit{PQFL} SGD update. The server aggregates the parameters of the local models and updates the global model parameters for the next round. }
    \label{fig: overview}
\end{figure}

\textcolor{black}{However, existing QFL approaches often assume uniform quantum circuit topologies and encoding methods across clients, an assumption that rarely holds in real-world quantum networks \cite{gurung2024personalized}. This heterogeneity, particularly in applications like anomaly detection, limits the effectiveness of naïve global model aggregation \cite{gurung2025performance}. To address this, we propose a PQFL architecture that allows each quantum client to train a customized local model while benefiting from federated knowledge exchange. PQFL is particularly suited for applications involving dispersed, sensitive, and context-specific data, such as industrial fault detection and privacy-preserving medical diagnostics \cite{gurung2025performance}. The overview of the proposed \textit{PQFL} method is shown in Fig.~\ref{fig: overview} and the comparison between our approach and other approaches is shown in Table~\ref{tab:fl_comparison}. By introducing customization to minimize client heterogeneity, PQFL addresses a fundamental gap in past QFL investigations, enabling robust anomaly detection in heterogeneous quantum networks \cite{chehimi2022quantum,wang2022data,ren2025toward}.}

\section{Proposed Methodology}

\subsection{Anomaly Detection Method}
This section presents an improved anomaly detection approach to solve two prevalent issues in real-world deployments: false detections (false positives) and missing detections (false negatives), introduced in work \cite{dong2024fadngs}.

\textit{False positives} arise when normal circumstances are incorrectly labeled as abnormalities.  These mistakes are frequently caused by overfitting to training data, inaccurately adjusted anomaly thresholds, or inherent data noise.  Client-side heterogeneity in QFL further exacerbates the problem, since differences in quantum encodings, circuit architecture, and local data distributions induce inconsistencies in model behavior.  When these locally biased quantum models are combined at the server, the final global model may provide unstable anomaly scores, resulting in higher false alarm rates.

\textit{False negatives} occur when the model fails to detect real abnormalities.  This is often caused by model underfitting, poor anomaly signals, or inadequate sensitivity to context-specific deviations.  In the QFL environment, heterogeneous clients may have incomplete or sparse representations of anomalous patterns, particularly for non-IID distributions.  Furthermore, quantum-specific noise sources such as decoherence, gate errors, and entanglement distortion can decrease the quality of feature embeddings, making it difficult to detect minor abnormalities from normal data.

\begin{figure*}
    \centering
    \includegraphics[width=0.79\linewidth]{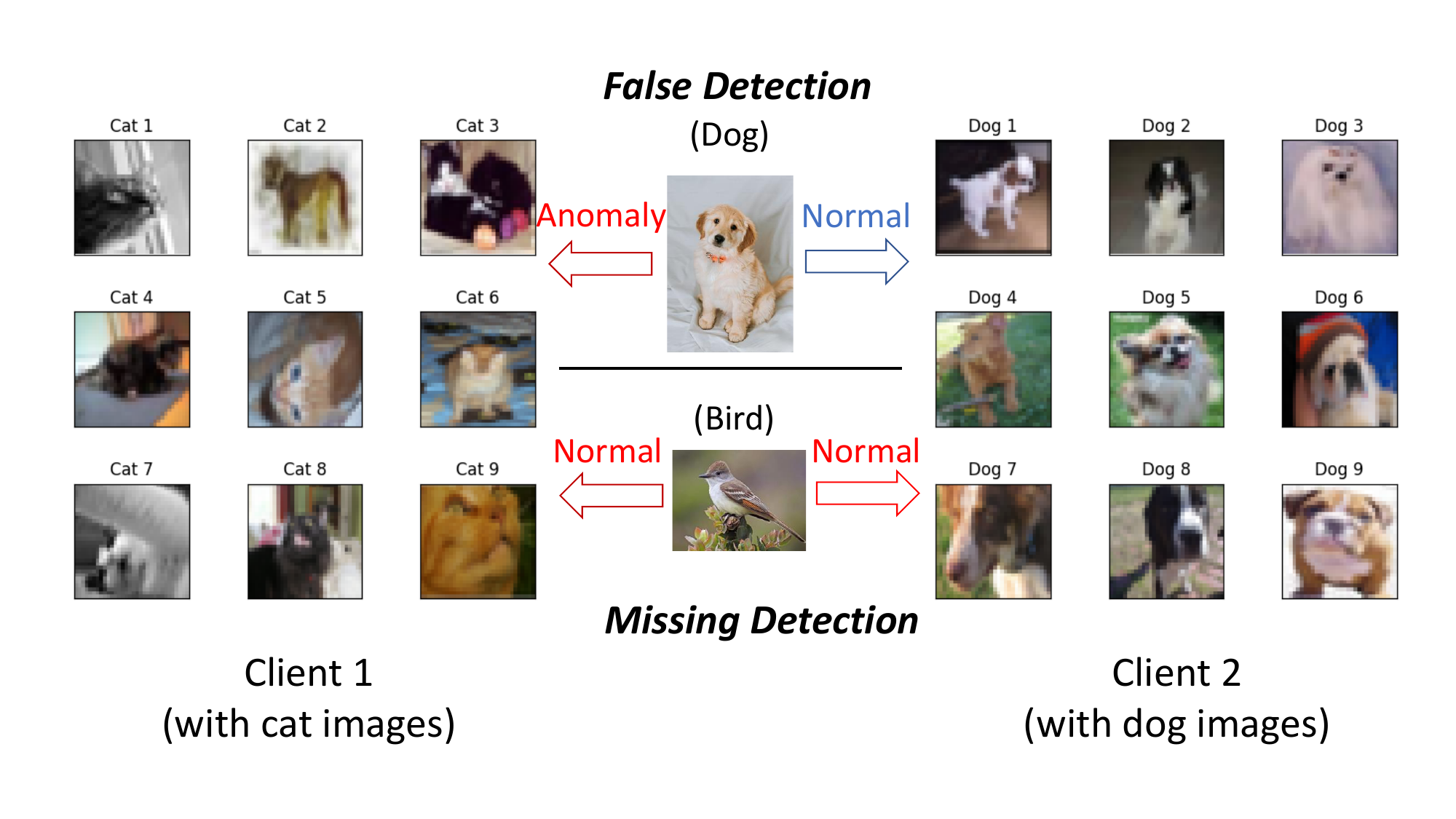}
    \vspace{-1em}
    \caption{An illustration of false and missing detection concerns in QFL anomaly detection.   In this scenario, photos of cats and dogs are considered normal, but those of other groups are considered abnormal. }
    \label{fig: anomaly}
\end{figure*}

To address these issues, our PQFL system implements personalized model updates that correspond to each client's quantum encoding approach and data properties.  PQFL decreases both types of errors while allowing cross-client knowledge transfer in realistic, noisy quantum settings.  Figure \ref{fig: anomaly} illustrates the method for detecting anomalies.

\subsection{QFL for Anomaly Detection}
\textbf{QML Model.} 
 We study a hybrid quantum-classical algorithm known as the variational quantum eigensolver, which is the leading computing method used in noisy intermediate-scale quantum (NISQ) devices. We use parameterized quantum circuits (PQC) to train the QML model, which is achieved by a classical optimization algorithm, such as stochastic gradient descent. 
 \textcolor{black}{Unlike traditional deep architectures, which simulate complex relations using large nonlinear layers, PQCs operate in exponentially vast Hilbert spaces, allowing for richer feature translations with fewer parameters. PQFL's higher-dimensional representation captures subtle connections between normal and anomalous samples, which conventional models cannot effectively convey. Thus, the PQC offers a true quantum advantage over nonlinearity by using superposition and entanglement to improve anomaly separability in non-IID federated situations.}
 For classification problems, our QNN model uses a PQC in a hybrid quantum-classical architecture. The input \(x \) is initially flattened and \textit{L2-normalized} as \( x = \frac{x}{|| x ||_2} \) ensuring a correct representation of the quantum state. The \textit{PQC} uses \(D_c \) qubits and comprises \(l \) layers. Each layer performs \textit{parameterized \(R_y \) rotations}, defined as \(R_y(\boldsymbol{w}) = \begin{bmatrix} \cos(\boldsymbol{w}/2) & -\sin(\boldsymbol{w}/2) \\ \sin(\boldsymbol{w}/2) & \cos(\boldsymbol{w}/2) \end{bmatrix} \). 
Next, \textit{CNOT (CX) gates} are used to introduce quantum entanglement. The trainable parameters, \(\boldsymbol{w} \in \mathbb{R}^{L \times n} \), are optimized to alter the input quantum state. Following quantum processing, we extract \textit{probabilities} 
\( p = |\psi|^2 \).
The final quantum state, represented by \(\psi \), is passed across a \textit{fully linked layer} \(y = Wp + b\) to map quantum-extracted features into the output space.  The details of our proposed method are elaborated via the following steps.

\textbf{Quantum Encoding.}
We provide a QFL algorithm that is specially made to identify lithography hotspots at pre-selected locations.  We prepare quantum states using the amplitude encoding method \cite{gonzalez2024efficient} for efficient modeling of quantum circuits.  The amplitudes of a \(n\)-qubit quantum state \(|\psi\rangle\) are encoded using this method as normalized classical data of \(N\) dimensions (\(N = 2^n\)), stated as 
\begin{equation}
    |\psi\rangle = \frac{1}{||x||} \sum_{j=1}^{N} x_j |j\rangle,
\end{equation}
where the state \(|\psi\rangle\) is normalized, and \(||x||\) indicates the norm of the vector \(x\).  Each dimension of the classical input data is indexed by \(j\) in this formula, which maps each component \(x_j\) to the associated quantum basis state \(|j\rangle\).

\textbf{Local QML model training.} We use VQE \cite{peruzzo2014variational} where the parameters defining a PQC's functioning are tuned using a classical optimizer based on PQC measurements.  The VQE tries to ensure that the output state of the PQC, denoted as \(|\Psi(\boldsymbol{w}_{n,k})| \), accurately approximates the ground state of a specified observable \(H \).  Mathematically, we can express it as
\begin{equation}
    \boldsymbol{w}_{n,k}^* = \arg\min_{\boldsymbol{w}_{n,k}} \langle H \rangle_{\Psi(\boldsymbol{w}_{n,k})},
\end{equation}
where the goal is to minimize the anticipated value of the observable (H) for the state \(|\Psi(\boldsymbol{w}_{n,k})|\).  For a suitably expressive a PQC, also known as an ansatz, the solution \(|\Psi(\boldsymbol{w}_{n,k}^*)| \) produced from this optimization closely approximates the desired eigenstate.  VQE applications include performing quadratic unconstrained binary optimization (QUBO) tasks. In VQE, the parameters \(\boldsymbol{w}_{n,k} \) of the PQC are often optimized iteratively using stochastic gradient descent (SGD).  SGD calculates the gradient \(\nabla_{\boldsymbol{w}_{n,k}} \langle H \rangle_{\Psi(\boldsymbol{w}_{n,k})} \) of the expectation value, which is determined from measurements of the PQC output.  This iterative approach allows for efficient convergence to the ideal values.

A noise-free PQC performs a unitary transformation $U(\boldsymbol{w}_{n,k})$ on \(n \) qubits, where \(\boldsymbol{w}_{n,k} = (\boldsymbol{w}_{n,k}^1, \dots, \boldsymbol{w}_{n,k}^D) \) is a vector with each component \(\boldsymbol{w}_{n,k}^d \in \mathbb{R} \) with \(d \in \mathcal{D} \). An ansatz often utilizes unitaries defined as
\begin{equation} \label{eq: unitary}
    U(\boldsymbol{w}_{n,k}) = \prod_{d=1}^{D} U_d(\boldsymbol{w}_{n,k}^d) V_d.
\end{equation}
Each unitary \(U_d(\boldsymbol{w}_{n,k}^d) \) is affected by the corresponding parameter \(\boldsymbol{w}_{n,k}^d \) in the vector \(\boldsymbol{w}_{n,k} \).  These gates are expressed as 
\begin{equation} \label{eq: unitary2}
    U_d(\boldsymbol{w}_{n,k}^d) = \exp \left( -i \frac{\boldsymbol{w}_{n,k}^d}{2} G_d \right),
\end{equation}
where \(G_d \) belongs to the collection \(\{I, X, Y, Z\}^{\otimes n} \), which represents the Pauli string generators.  The sequence's unitary \(V_d \) is static and does not rely on parameters.

Initially in the base state \( |0\rangle \), the pristine PQC acts on \( n \) qubits to produce the quantum state
\begin{equation} \label{eq: state}
        | \Psi(\boldsymbol{w}_{n,k})\rangle = U(\boldsymbol{w}_{n,k})|0\rangle.
\end{equation}
The following, therefore, provides the relevant density matrix for this state
\begin{equation} \label{eq: state2}
    \Psi(\boldsymbol{w}_{n,k}) = | \Psi(\boldsymbol{w}_{n,k})\rangle \langle \Psi(\boldsymbol{w}_{n,k}) |.
\end{equation}

In order to solve the following optimization problem, the Variational Quantum Eigensolver (VQE) seeks to determine the ideal parameters \( \boldsymbol{w}_{n,k} \) as
\begin{equation} \label{eq:problem1}
        \min_{\boldsymbol{w}_{n,k} \in \mathbb{R}^D} \{ f_n(\boldsymbol{w}_{n,k}) := \langle H \rangle _{|\Psi(\boldsymbol{w}_{n,k})\rangle} \},
\end{equation}
where the anticipated value is represented by the loss function \( f_n(\boldsymbol{w}_{n,k}) \) and with relation to an observable 
\begin{equation}
    \langle H \rangle _{|\Psi(\boldsymbol{w}_{n,k})\rangle} = \operatorname{Tr}(H \Psi(\boldsymbol{w}_{n,k})),
\end{equation}
where $\operatorname{Tr}$ is expressed as a trace of a matrix.  The projection operators onto the eigenspaces of \( H \) can also be used to express this loss function
\begin{equation}
       f_n(\boldsymbol{w}_{n,k}) = \sum_{y=1}^{N_h} h_y  \operatorname{Tr}(\Pi_y \Psi(\boldsymbol{w}_{n,k})).
\end{equation}
The optimal loss value is denoted as \( f_n^* \) can be described as
\begin{equation}
     f_n^{*}=\min_{\boldsymbol{w} \in \mathbb{R}^D} f_{n,k}(\boldsymbol{w}_{n,k}).
\end{equation}

The eigendecomposition of the observable defined by the Hermitian matrix \( H \) may be expressed as 
\begin{equation} \label{eq: Hermitian}
    H = \sum_{y=1}^{N_h} h_y \Pi_y,
\end{equation}
where \( N_h \leq 2^n \) and \( \{\Pi_y\}_{y=1}^{N_h} \) are the projection operators for the corresponding eigenspaces, and \( \{h_y\}_{y=1}^{N_h} \) are the distinctive eigenvalues of \( H \).  This makes it possible to express the loss function \( f_n(\boldsymbol{w}_{n,k}) \) in the particular problem as
\begin{equation} \label{eq: sloss}
    f_n(\boldsymbol{w}_{n,k}) = \sum_{y=1}^{N_h} h_y \operatorname{Tr}(\Pi_y \Psi(\boldsymbol{w}_{n,k})).
\end{equation}

The VQE algorithm is shown in Algorithm~\ref{algo:vqe}. VQE method optimizes the parameters of a parameterized quantum circuit (PQC) in order to estimate the ground state energy of an observable $H$. It starts with $H$, the number of iterations $T$, the learning rate $\eta$, and starting parameters $\boldsymbol{w}_{n,k}^{(0)}$, then initializes the iteration counter $t=0$ and sets $\boldsymbol{w}_{n,k} = \boldsymbol{w}_{n,k}^{(0)}$ (Line 1). Each iteration from $t=1$ to $T$ (Line 2) builds the quantum state $|\Psi(\boldsymbol{w}_{n,k})\rangle = U(\boldsymbol{w}_{n,k}) |0\rangle$ using the PQC unitary. The expectation value $f_n(\boldsymbol{w}_{n,k}) = \langle H \rangle_{|\Psi(\boldsymbol{w}_{n,k})\rangle} = \operatorname{Tr}(H \Psi(\boldsymbol{w}_{n,k}))$ (Line 4), computes the gradient $\nabla f_n(\boldsymbol{w}_{n,k})$ (Line 5), and updates the parameters using stochastic gradient descent as $\boldsymbol{w}_{n,k} \leftarrow \boldsymbol{w}_{n,k} - \eta \nabla f_n(\boldsymbol{w}_{n,k})$ (Line 6).  Following $T$ iterations (Line 7), it provides the optimized parameters $\boldsymbol{w}_{n,k}^*$ and the approximated ground state energy $f_n(\boldsymbol{w}_{n,k}^*)$ (Line 8).
\begin{algorithm}[t]
\caption{Variational Quantum Eigensolver (VQE)}
\begin{algorithmic}[1]
\REQUIRE Observable $H$, number of iterations $T$, learning rate $\eta$, initial parameters $\boldsymbol{w}_{n,k}^{(0)}$
\STATE Set $t = 0$, $\boldsymbol{w}_{n,k} = \boldsymbol{w}_{n,k}^{(0)}$
\FOR{$t = 1$ to $T$}
\STATE Prepare the quantum state $|\Psi(\boldsymbol{w}_{n,k})\rangle = U(\boldsymbol{w}_{n,k}) |0\rangle$
\STATE Measure the expectation value $f_n(\boldsymbol{w}_{n,k}) = \langle H \rangle_{|\Psi(\boldsymbol{w}_{n,k})\rangle}$
\STATE Compute the gradient $\nabla f_n(\boldsymbol{w}_{n,k})$
\STATE Update parameters: $\boldsymbol{w}_{n,k} \leftarrow \boldsymbol{w}_{n,k} - \eta \nabla f_n(\boldsymbol{w}_{n,k})$
\ENDFOR
\STATE \textbf{Output:} Optimized parameters $\boldsymbol{w}_{n,k}^* = \boldsymbol{w}_{n,k}$, approximated ground state energy $f_n(\boldsymbol{w}_{n,k}^*)$
\end{algorithmic}
\label{algo:vqe}
\end{algorithm}

\textbf{Model Aggregation for QFL.} The QFL framework uses the VQE method to compute the local loss functions $f_n(\boldsymbol{w}_{n,k})$ and optimize the local parameters $\boldsymbol{w}_{n,k}^*$ for each client $n \in N_c$. A global model is then produced by aggregating the locally optimized parameters.  The aggregation procedure aggregates the parameters from all $N_c$ clients to create a global parameter set $\boldsymbol{w}_{\text{global}}$, which is then utilized to update the PQC for the next cycle of training.  A basic but effective aggregation approach is to compute the weighted average of the local parameters, where the weights $\alpha_n$ reflect the contribution of each client (for example, depending on data size or model performance). The global parameters are obtained as
\begin{equation} \label{eq:aggregation}
\boldsymbol{w}_{\text{global}} = \frac{1}{N} \sum_{n=1}^{N_c}  \boldsymbol{w}_{n,k}^*,
\end{equation}
where $\boldsymbol{w}_{n,k}^*$ represents the optimized parameter from the $n$-th client. This aggregated parameter set is then distributed to all clients to start the next round of local VQE training, allowing for collaborative learning.

\begin{algorithm}[t]
\caption{Quantum Federated Learning (QFL)}
\begin{algorithmic}[1]
\REQUIRE Number of clients $N_c$, number of rounds $R$, initial global parameters $\boldsymbol{w}_{\text{global}}^{(0)}$, weights $\{\alpha_n\}_{n=1}^{N_c}$ with $\sum_{n=1}^{N_c} \alpha_n = 1$
\STATE Initialize $\boldsymbol{w}_{\text{global}} = \boldsymbol{w}_{\text{global}}^{(0)}$
\FOR{$r = 1$ to $R$}
\FOR{each client $n = 1$ to $N_c$ in parallel}
\STATE Set local parameters $\boldsymbol{w}_{n,k} = \boldsymbol{w}_{\text{global}}$
\STATE Compute VQE to obtain optimized local parameters $\boldsymbol{w}_{n,k}^*$ and loss $\hat{f_n}(\boldsymbol{w}_{n,k}^*)$
\STATE Send $\boldsymbol{w}_{n,k}^*$ to server
\ENDFOR
\STATE Aggregate parameters: $\boldsymbol{w}_{\text{global}} = \frac{1}{N} \sum_{n=1}^{N_c}  \boldsymbol{w}_{n,k}^*,
$
\STATE Broadcast $\boldsymbol{w}_{\text{global}}$ to all clients
\ENDFOR
\STATE \textbf{Output:} Global parameters $\boldsymbol{w}_{\text{global}}^* = \boldsymbol{w}_{\text{global}}$
\end{algorithmic}
\label{algo: qfl}
\end{algorithm}

The QFL algorithm shown in Algorithm~\ref{algo: qfl} coordinates numerous clients to train a global quantum model with VQE.  It accepts as input the number of clients $N_c$, number of rounds $R$, initial global parameters $\boldsymbol{w}_{\text{global}}^{(0)}$, and client weights $\{\alpha_n\}_{n=1}^{N_c}$ with $\sum_{n=1}^{N_c} \alpha_n = 1$, and then initializes the global parameters.  Line 2: $\boldsymbol{w}_{\text{global}} = \boldsymbol{w}_{\text{global}}^{(0)}$.  For each round $r$ from 1 to $R$ (Line 2), each client $n$ from 1 to $N_c$ concurrently (Line 3) sets its local parameters to global parameters.  $\boldsymbol{w}_{n,k} = \boldsymbol{w}_{\text{global}}$ (Line 4), calculates VQE to optimize local parameters $\boldsymbol{w}_{n,k}^*$, obtains the empirical loss $\hat{f_n}(\boldsymbol{w}_{n,k}^*)$ (Line 5), and transmits the improved parameters to the server (Line 6).  The server computes an average $\boldsymbol{w}_{\text{global}} = \frac{1}{N} \sum_{n=1}^{N_c}  \boldsymbol{w}_{n,k}^*$ (Line 8) and sends the revised global parameters to all clients (Line 9).  After $R$ rounds (Line 10), the algorithm returns the final global parameters $\boldsymbol{w}_{\text{global}}^*$ (Line 11), allowing for collaborative quantum model training.

\subsection{Proposed PQFL method for Anomaly Detection} 

\textbf{Model Personalization for QML model training.}
\textcolor{black}{In our paper, we study heterogeneous quantum encodings across clients, i.e., clients employ distinct quantum encoding strategies to encode their local classical data, such as basis encoding, amplitude encoding, or phase encoding. This variance leads to inconsistencies in quantum state representations, resulting in differences in feature extraction, model updates, and training dynamics between clients. As a result, heterogeneous encoding adds new challenges to QFL, influencing model aggregation, convergence stability, and generalization.
If all clients employ the same encoding approach, there are still differences between them due to data preprocessing, feature scaling, and quantum circuit implementation.  These modifications result in separate quantum state representations, even when the same encoding approach is used, resulting in client-specific quantum data distributions.  This incorporates encoding-induced heterogeneity, which affects model training, aggregation, and convergence in QFL.}

This section focuses on utilizing SGD to create the VQE. The SGD technique iteratively updates the parameter vector $\boldsymbol{w}_{n,k}$, starting with an initial parameter set $\boldsymbol{w}_{n,k}^{0} \in \mathbb{R}^D$ as 
\begin{equation}
        \boldsymbol{w}_{n,k}^{t+1} = \boldsymbol{w}_{n,k}^t - \eta_t \left( \hat{g}_{n,k}^t + \lambda (\bm{w}_{n,k}^{t} - \bm{w}_k) \right).
\end{equation}
where the gradient $\hat{g}_{n,k}^t$ is given as
\begin{equation} \label{eq: eloss}
\hat{g}_{n,k}^t = \nabla f_n(\boldsymbol{w}_{n,k})|_{\boldsymbol{w}_{n,k}=\boldsymbol{w}_{n,k}^t} = \begin{bmatrix}
\frac{\partial f_n(\boldsymbol{w}_{n,k})}{\partial \boldsymbol{w}_{n,k}^1} \\
\vdots \\
\frac{\partial f_n(\boldsymbol{w}_{n,k})}{\partial \boldsymbol{w}_{n,k}^D}
\end{bmatrix}.
\end{equation}

\textcolor{black}{The intuition behind this personalized learning approach is to balance local quantum variations caused by heterogeneous quantum encodings across different clients while maintaining alignment with the global model. The regularization term \( \lambda (\bm{w}_{n,k}^{t} - \bm{w}_k) \) plays a crucial role in mitigating inconsistencies that arise when different clients use distinct quantum encoding schemes, such as amplitude encoding, phase encoding, or basis encoding, which lead to non-uniform feature representations. Directly averaging updates from such heterogeneous encodings can degrade model performance, as quantum parameters may not be directly comparable. By softly guiding local models toward a shared global representation while preserving the flexibility to adapt to individual encoding methods, this approach enhances global model generalization with respect to model personalization. The tuning parameter \( \lambda \) determines the strength of this regularization, balancing personalization and generalization—where a higher \( \lambda \) enforces stronger alignment with the global model, and a lower \( \lambda \) allows for greater local adaptation based on each client’s quantum-specific constraints.}

\textbf{Global model aggregation.} To compile and produce the new global model, the central server gathers model updates from every client using Eq.~\eqref{eq:aggregation}.
This global aggregate produces a more adaptive global model, which accounts for any customized updates from quantum clients that finish more epochs. While still contributing to the global model, the proposed personalized approach allows each client to effectively optimize its unique data distribution. By dynamically adjusting the number of local epochs according to the performance of the global model, \textit{PQFL} aims to improve both overall convergence and individual client performance.

 \subsection{Proposed  Algorithm} 
 \begin{algorithm}[t]
\caption{Personalized Quantum Federated Learning (PQFL)}
\begin{algorithmic}[1]
\REQUIRE Number of clients $N_c$, global rounds $K$, local epochs $T$, initial global parameters $\boldsymbol{w}_{\text{global}}^{(0)}$, weights $\{\alpha_n\}_{n=1}^{N_c}$ with $\sum_{n=1}^{N_c} \alpha_n = 1$, learning rate $\eta$, regularization parameter $\lambda$
\STATE Initialize $\boldsymbol{w}_{\text{global}} = \boldsymbol{w}_{\text{global}}^{(0)}$ \label{line2}
\FOR{each global round $k = 1$ to $K$} \label{line3}
\STATE Broadcast $\boldsymbol{w}_{\text{global}}$ to all clients \label{line4}
\FOR{each client $n = 1$ to $N_c$ in parallel} \label{line5}
\STATE Set local parameters $\boldsymbol{w}_{n,k} = \boldsymbol{w}_{\text{global}}$ \label{line6}
\STATE Compute VQE and loss estimation to obtain $\boldsymbol{w}_{n,k}^*$ and $\hat{f_n}(\boldsymbol{w}_{n,k}^t)$ using Algorithms 1 and 2, with update rule: $\boldsymbol{w}_{n,k}^{t+1} = \boldsymbol{w}_{n,k}^t - \eta \left[ \hat{g}_{n,k}^t + \lambda (\boldsymbol{w}_{n,k}^t - \boldsymbol{w}_{\text{global}}) \right]$ \label{line7}
\STATE Send $\boldsymbol{w}_{n,k}^*$ to server \label{line8}
\ENDFOR \label{line9}
\STATE Aggregate parameters: $\boldsymbol{w}_{\text{global}} = \sum_{n=1}^{N_c} \alpha_n \boldsymbol{w}_{n,k}^*$ \label{line10}
\STATE Evaluate global model $\boldsymbol{w}_{\text{global}}$ via validation test \label{line11}
\STATE Broadcast $\boldsymbol{w}_{\text{global}}$ to all clients \label{line12}
\ENDFOR \label{line13}
\STATE \textbf{Output:} Optimal global parameters $\boldsymbol{w}_{\text{global}}^*$ \label{line14}
\STATE Compute anomaly score for each data point based on deviation or reconstruction error \label{line15}
\end{algorithmic}
\label{algo:my_algo}
\end{algorithm}

We summarize our proposed \textit{PQFL} methodology in Algorithm~\ref{algo:my_algo}. The PQFL approach enables collaborative training of a quantum model across $N_c$ clients while providing personalized updates. It accepts as input the number of clients $N_c$, global rounds $K$, local epochs $T$, and initial global parameters $\boldsymbol{w}_{\text{global}}^{(0)}$, weights $\{\alpha_n\}_{n=1}^{N_c}$ with $\sum_{n=1}^{N_c} \alpha_n = 1$, learning rate $\eta$, and regularization parameter $\lambda$, then initializes $\boldsymbol{w}_{\text{global}} = \boldsymbol{w}_{\text{global}}^{(0)}$. For each global round $k$ from 1 to $K$ (Line 2), the server broadcasts $\boldsymbol{w}_{\text{global}}$ to all clients (Line 3). Each client $n$ from 1 to $N_c$ in parallel (Line 4) provides local parameters. Algorithms~\ref{algo:vqe} are used to compute VQE, with parameters being updated using a customizable rule to retrieve $\boldsymbol{w}_{n,k}^*$ and $\hat{f_n}(\boldsymbol{w}_{n,k}^t)$ (Line 6), and transmit $\boldsymbol{w}_{n,k}^*$ to the server (Line 7).  The server aggregates parameters as $\boldsymbol{w}_{\text{global}} = \sum_{n=1}^{N_c} \alpha_n \boldsymbol{w}_{n,k}^*$ (Line 9), validates the global model (Line 10), and sends $\boldsymbol{w}_{\text{global}}$ to clients (Line 11).  Following $K$ rounds (Line 12), it returns the ideal global parameters $\boldsymbol{w}_{\text{global}}^*$ (Line 13) and computes anomaly scores based on deviation or reconstruction error (Line 14).

\subsection{Computational Complexity}
{\color{black}
\textbf{Time Complexity:} Time complexity analysis of the suggested \textit{PQFL} algorithm can be decomposed by looking at the nested loops in Algorithm~\ref{algo:my_algo}.  Let \(N\) be the total number of clients that participate, \(K\) be the number of global rounds, and \(T\) be the number of local training epochs.  Each client (\(N\) in parallel) executes \(T\) local training epochs during each global round (run \(K\).  Each global round has a total work of \(N \times T \times \mathcal{O}(\text{GradEval})\), assuming that \(\mathcal{O}(\text{GradEval})\) is the time to evaluate gradients or parameter updates on one client for a single epoch (including forward and backward passes based on quantum-circuits). The entire time complexity across \(K\) global rounds is therefore \(\mathcal{O}(K \, N \, T \times \text{GradEval})\). 
\[\text{Time complexity} = \mathcal{O}(K \, N \, T \times \text{GradEval}).\]

\textbf{Space Complexity:} Each client must keep the local model parameters for memory usage, which, assuming that the model contains \(p\) trainable parameters, is \(\mathcal{O}(p)\).  In order to compute gradients or parameter shifts, clients also store intermediate variables, usually of the same order \(\mathcal{O}(p)\). A global model of size \(\mathcal{O}(p)\) is stored on the server side, and temporary storage \(\mathcal{O}(N \, p)\) can be needed when aggregating updated parameters from the \(N\) clients.  Therefore, while taking into account all clients and the server collectively, the total space complexity is \(\mathcal{O}(N \, p)\). In every round, when aggregation is finished, the server's final storage is reduced to \(\mathcal{O}(p)\) to store the updated global model.
\[\text{Space complexity} = \mathcal{O}(N \, p).\]
}


\section{Simulations and Evaluation}
In this section, we discussed our datasets, simulation results, and their outcomes. For comparison, we have provided an extensive set of simulation results to compare with various methods for non-IID data distribution in  Table~\ref{tab:data_distribution_niid}.

\subsection{Datasets}
We employ CIFAR-10 \cite{krizhevsky2009learning} as the baseline dataset to reflect the normal data settings for anomaly detection.  This dataset contains a collection of images from ten classes that are diverse but consistently labeled.   This dataset's broad applicability and well-established standards provide a solid foundation for evaluating normalcy models. For anomaly data, we employ datasets with different class distributions and levels of complexity, including ImageNet \cite{deng2009imagenet}, SVHN \cite{netzer2011reading}, and CIFAR-100 \cite{krizhevsky2009learning} because of their depth and complexity. CIFAR-100, with its 100 classes, offers greater diversity. SVHN, which contains real-world digit images, differs significantly from the CIFAR collection. ImageNet, which began with 64x64 pixel images, has been downsampled to 32x32 to match the resolution of the other datasets, increasing the detection challenge by challenging the model to locate anomalies in lower-quality images. Our simulations predominantly use CIFAR-10 for normal data and CIFAR-100 for anomalies, allowing for a detailed study of model performance under a variety of challenging conditions.

\textcolor{black}{The standard CV benchmarks CIFAR-10/100, SVHN, and ImageNet subsets lack any provable, inherent quantum advantage.  Instead of any quantum-favorable pixel structure, our non-IID architecture (avg.\ pairwise KL divergence $\approx 0.8$) mainly stresses cross-client heterogeneity.  Therefore, rather than a dataset-inherent quantum shortcut, we understand PQFL's benefits as resulting from the synergy of quantum feature mappings with federated personalization under heterogeneity.}

\begin{figure*}
    \centering
    \includegraphics[width=0.90\linewidth]{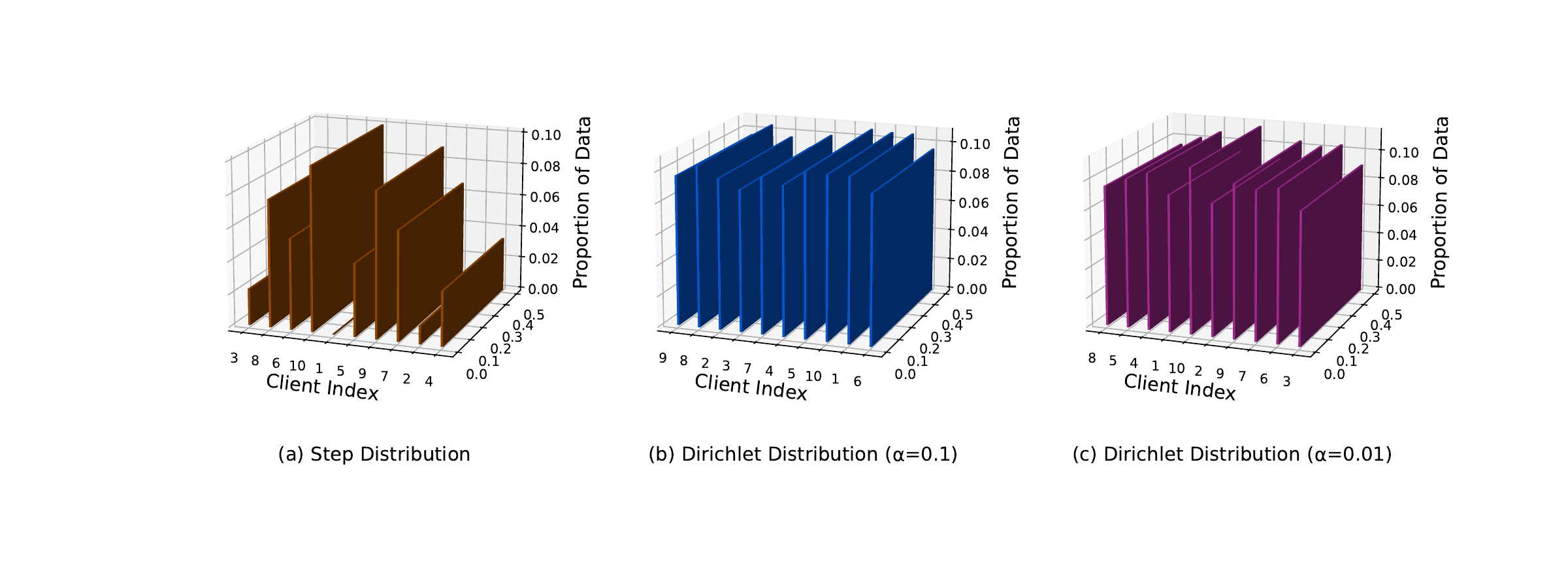}
    \vspace{-3em}
    \caption{Non-IID data distribution with three types of distribution. (a) is 'Step' data distribution and (b) and (c) are both 'Dirichlet' data distribution with learning rates of 0.1 and 0.01, respectively, across 10 clients. The figure demonstrates that the step is more skewed than Dirichlet and among Dirichlet, the parameter of $\alpha = 0.01$ is more skewed. }
    \label{fig:non-iid}
\end{figure*}

\subsection{Simulation Settings} 
We present a QFL system that trains a QNN for image classification using ten quantum clients and a single quantum server.  This setup is supported by the \texttt{torchquantum} library, which makes output measurements easier and converts quantum states to classical bits for classification. In most of the simulation results, we have used ten clients where each client follows the VQE, where the clients use the QNN model and update their local model using regularization based \textit{PQFL} using the classical SGD.

\textbf{Non-IID data distribution.} We consider $N_c = 10$ clients, each with a local dataset exhibiting non-identical and independently distributed (non-IID) characteristics, in order to simulate the effects of heterogeneous data in the QFL framework.  The non-IID character stems from differences in the classical input data $x_n$ for each client $n$, representing genuine anomaly detection settings in which clients have distinct distributions of normal and anomalous data points.  Each client's dataset has a unique label distribution, with the fraction of anomalous samples varying from 5\% to 50\% among clients. This leads to skewed feature spaces following amplitude encoding into quantum states. To quantify this heterogeneity, we have calculated the Kullback-Leibler (KL) divergence between client feature distributions, and have found an average pairwise KL divergence of 0.8, suggesting strong data heterogeneity.  This heterogeneity affects the local loss functions $\hat{f_n}(\boldsymbol{w}_{n,k})$ produced by method 2, altering training stability and demanding individualized modifications in the PQFL method (Algorithm 3), balancing the performance of the local and global model. We have also considered the Dirichlet distributions of ($\alpha = 0.1$ and $\alpha = 0.01$) for non-IID comparison as shown in Fig.~\ref{fig:non-iid}. 

\textbf{Parameter settings.} The PQFL framework for anomaly detection is developed using the Qiskit library (version 1.1.0) and the Qiskit FGDAEr simulator (version 0.14.2) on a Linux operating system (Ubuntu 22.04 LTS) powered by an NVIDIA RTX 4090 GPU with 24 GB of VRAM to speed quantum circuit simulations.  The experimental setup includes \(N_c = 10 \) clients, each training a PQC with \(D_c = 4 \) qubits and \(l = 3 \) layers, and employing amplitude encoding for quantum state preparation.  The number of global communication rounds is set to \(K = 50 \), and each client executes 20 local training epochs each round. To balance local and global model alignment, the regularization parameter for personalized updates in PQFL is \( \lambda = 0.1 \), whereas the learning rate for stochastic gradient descent is \( \eta = 0.01 \).  Empirical loss estimation requires 1000 measurements to accurately approximate the expected value \(\hat{f}_n(\boldsymbol{w}_{n,k}) \).  Client weights \(\{\alpha_n\}_{n=1}^{N_c} \) are allocated depending on dataset sizes, normalized so that \(\sum_{n=1}^{N_c} \alpha_n = 1 \), with \(\alpha_n \in [0.05, 0.15] \) representing variable local dataset sizes (ranging from 500 to 2000 samples per client).  This setup takes advantage of the NVIDIA RTX 4090 GPU's computing capabilities to efficiently conduct quantum circuit simulations and gradient calculations in the Qiskit FGDAEr environment.

\textbf{Performance matrices.} 
To evaluate the performance of our approach, we use \textit{False Error} (FE), also known as false positive rate, \textit{Missing Error} (ME), also known as false negative rate to calculate the loss value. We can define FE and ME as
\begin{equation}
    \text{FE } (\%) = \frac{\text{FP}}{\text{TP}+\text{FP}} *100,
\end{equation}
\begin{equation}
    \text{ME } (\%) = \frac{\text{FN}}{\text{TP}+\text{FN}} *100,
\end{equation}
where FP is false-positive, denotes normal dataset samples predicted as anomaly samples, FN is false-negative, denotes dataset samples that are anomaly but detected as normal, and TP is true-positive, denotes a data sample is an anomaly and also predicted as an anomaly. Note that the lower value of FE and ME represents better performance. We also use \textit{Area Under the Receiver Operating Characteristic Curve} (AUROC) \cite{bradley1997use} and \textit{Area Under the Precision-Recall Curve} (AUPR) \cite{boyd2013area} matrices for evaluation, where AUROC assesses the model's ability to discriminate between normal and anomaly cases under various threshold settings and AUPR examines the trade-off between precision and recall, concentrating mainly on the model's performance in handling imbalanced datasets with few anomalies. Note that higher AUROC and AUPR scores indicate better performance. 
\begin{figure}
    \centering
    \footnotesize
    \begin{subfigure}[t]{0.49\linewidth} 
        \centering
        \includegraphics[width=\linewidth]{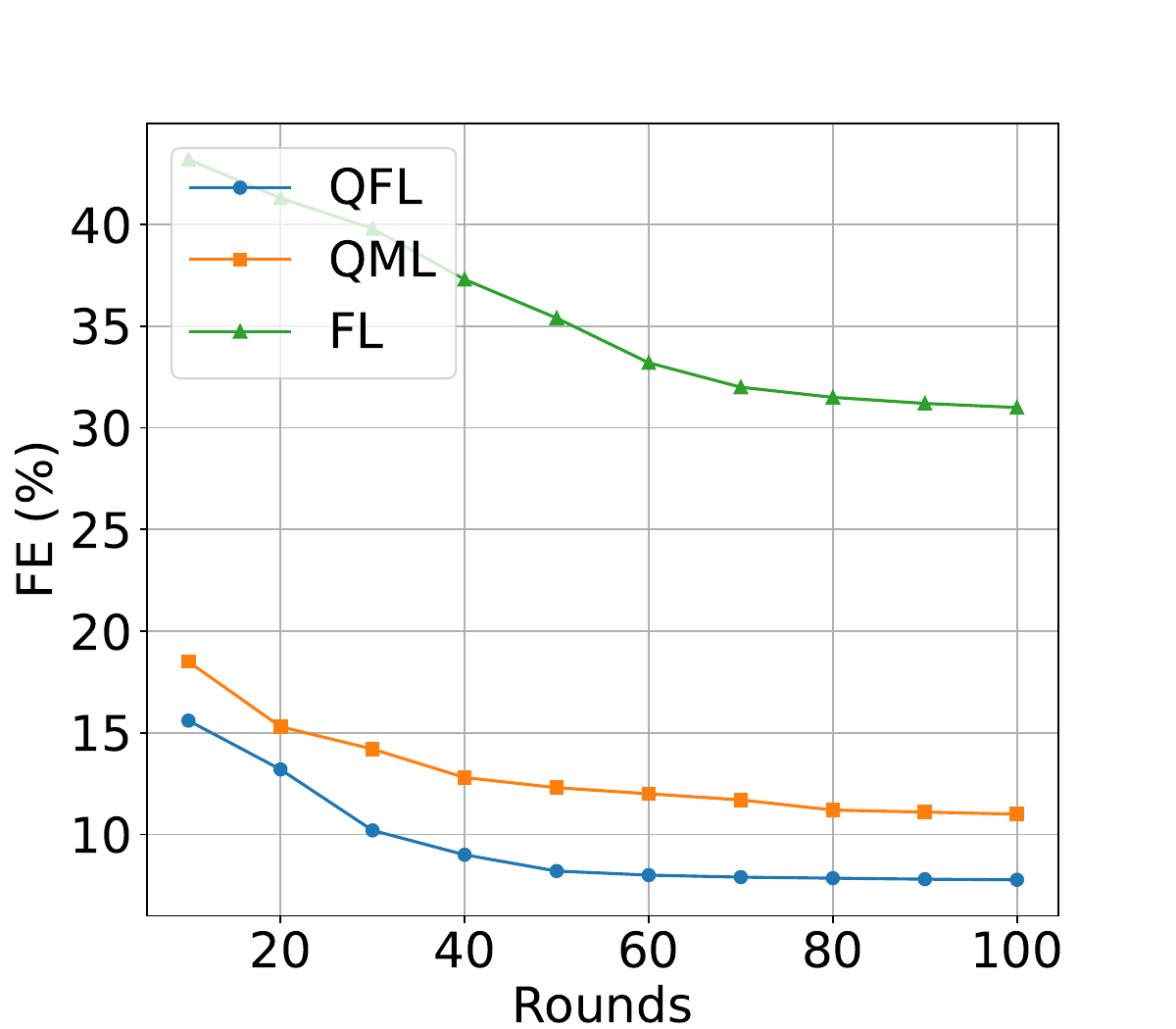}
        \caption{FE (\%)}
        \label{fig1a}
    \end{subfigure}
    \hfill 
    \begin{subfigure}[t]{0.49\linewidth} 
        \centering
        \includegraphics[width=\linewidth]{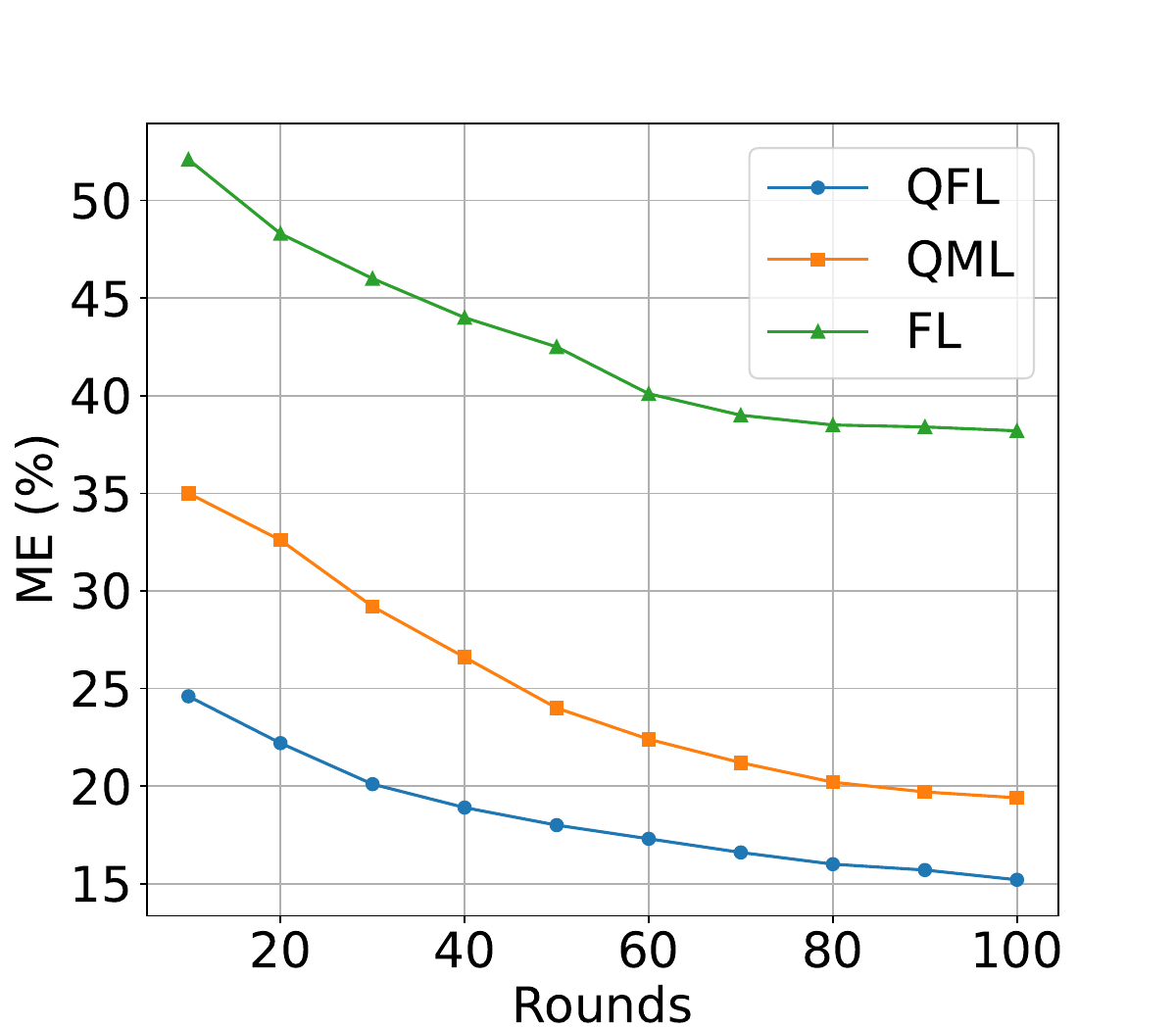}
        \caption{ME (\%)}
        \label{fig1b}
    \end{subfigure}
    \caption{Performance comparison between QFL, QNN, and FL approaches. we compared both FE (\%) (a) and ME (\%) (b) in anomaly detection where we use CIFAR-10 as normal dataset and CIFAR-100 as anomaly dataset. }
    \label{fig1}
\end{figure}
\begin{table}
    \centering
        \caption{Ablation experiments on qubits $D_q$ in QFL. Qubits used for this comparison are 2,3,5, and 10. CIFAR-10 can not run on 1 qubit as it is a high-dimensional image data (32 x 32 x 3 pixels) with complex figures. In addition, we have decided against adding more qubits because 10 qubits demand significant processing power and resources.}
    \label{tab: qubits}
    \begin{tabular}{p{0.25cm}|p{1.4cm}|p{1.4cm}|p{2cm}|p{1.8cm}}
        \hline
        $D_q$ & FE(\%) $\downarrow$ & ME(\%) $\downarrow$ & AUROC(\%) $\uparrow$ & AUPR(\%) $\uparrow$ \\
        \hline
        2  & 31.90$\pm$1.3  & 52.60$\pm$0.7  & 80.2$\pm$0.6 & 80.5$\pm$0.6  \\
        3  & 26.2$\pm$0.8  & 46.9$\pm$1.5  & 81.8$\pm$0.9  & 82.0$\pm$2.1  \\
        5  & 12.3$\pm$0.6  & 32.0$\pm$0.7  & 83.5$\pm$0.7  & 86.8$\pm$0.9  \\
        10 &  \textbf{8.29$\pm$1.4} & \textbf{25.72$\pm$0.6} & \textbf{84.2$\pm$0.9} & \textbf{91.88$\pm$0.9} \\ 
        \hline
    \end{tabular}
\end{table}
\begin{table}
    \centering
        \caption{Ablation experiments on quantum layers $l$ in QFL. For comparison, we use layers 1, 2, 3, 5, and 10. Fewer quantum layers result in underfitting with low expressiveness, whereas more layers might improve representation but risk barren plateaus, hindering efficient training.}
    \label{tab: qlayer}
    \begin{tabular}{c|c|c|c|c}
        \hline
        $l$ & FE(\%) $\downarrow$ & ME(\%) $\downarrow$ & AUROC(\%) $\uparrow$ & AUPR(\%) $\uparrow$ \\
        \hline
        1  & \textbf{8.29$\pm$1.4}  & \textbf{25.4$\pm$1.0}  & \textbf{84.6$\pm$0.6} & \textbf{92.1$\pm$2.1}  \\
        2  & 8.37$\pm$1.2  & 26.2$\pm$0.7  & 78.2$\pm$0.6 & 90.5$\pm$1.6  \\
        3  & 8.60$\pm$0.7  & 25.9$\pm$1.0  & 82.3$\pm$1.9  & 92.0$\pm$2.1  \\
        5  & 9.01$\pm$0.6  & 27.3$\pm$0.8  & 82.7$\pm$2.7  & 91.8$\pm$1.0  \\
        10 & 8.56$\pm$2.1  & 26.6$\pm$1.1  & 81.4$\pm$1.7  & 91.8$\pm$0.9  \\ 
        \hline
    \end{tabular}
\end{table}
\begin{figure*}[ht]
    \centering
    \begin{subfigure}{0.247\textwidth}
        \includegraphics[width=\linewidth]{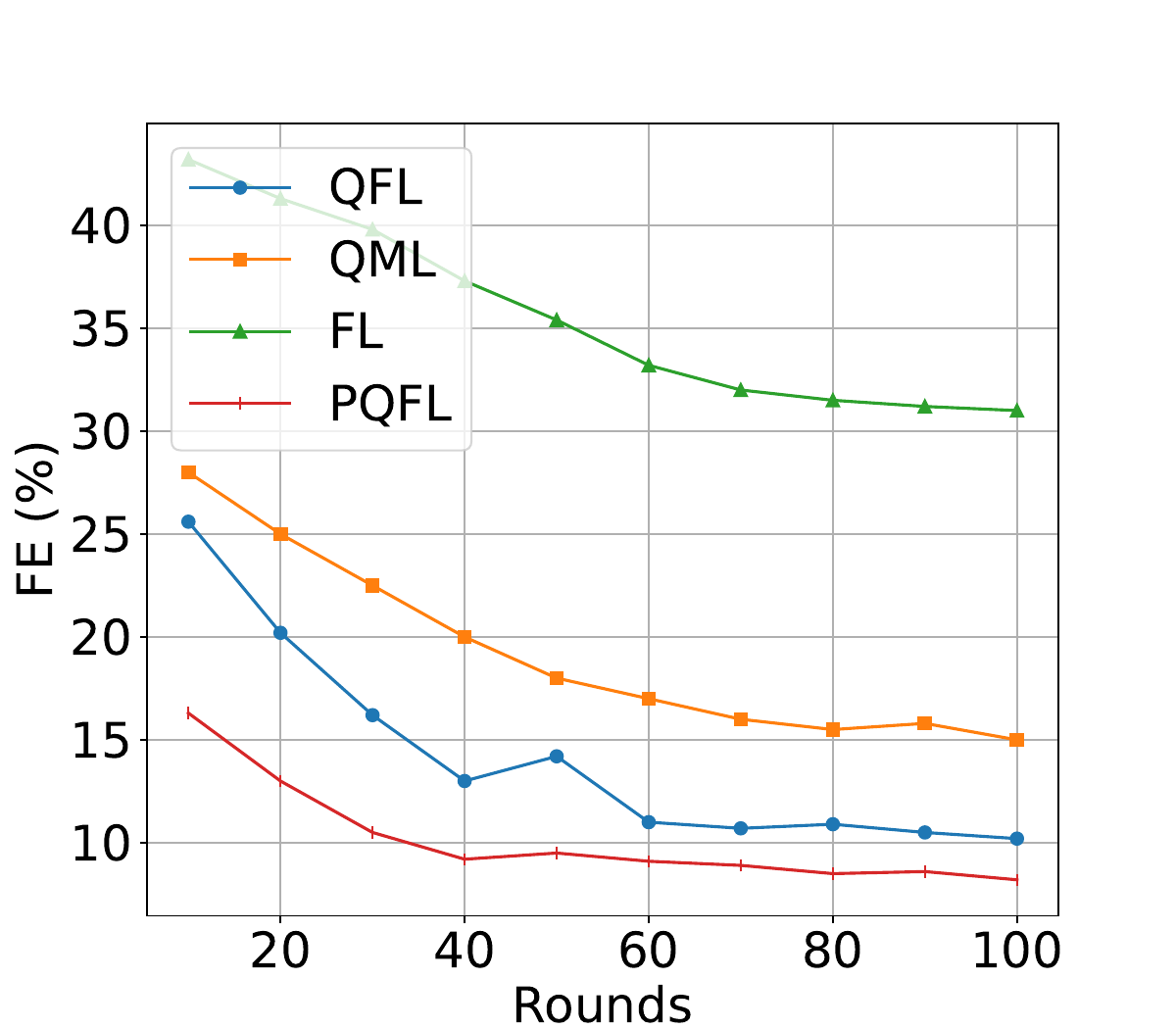}
        \caption{False Error}
        \label{fig:nplot1}
    \end{subfigure}%
    \hfill
    \begin{subfigure}{0.247\textwidth}
        \includegraphics[width=\linewidth]{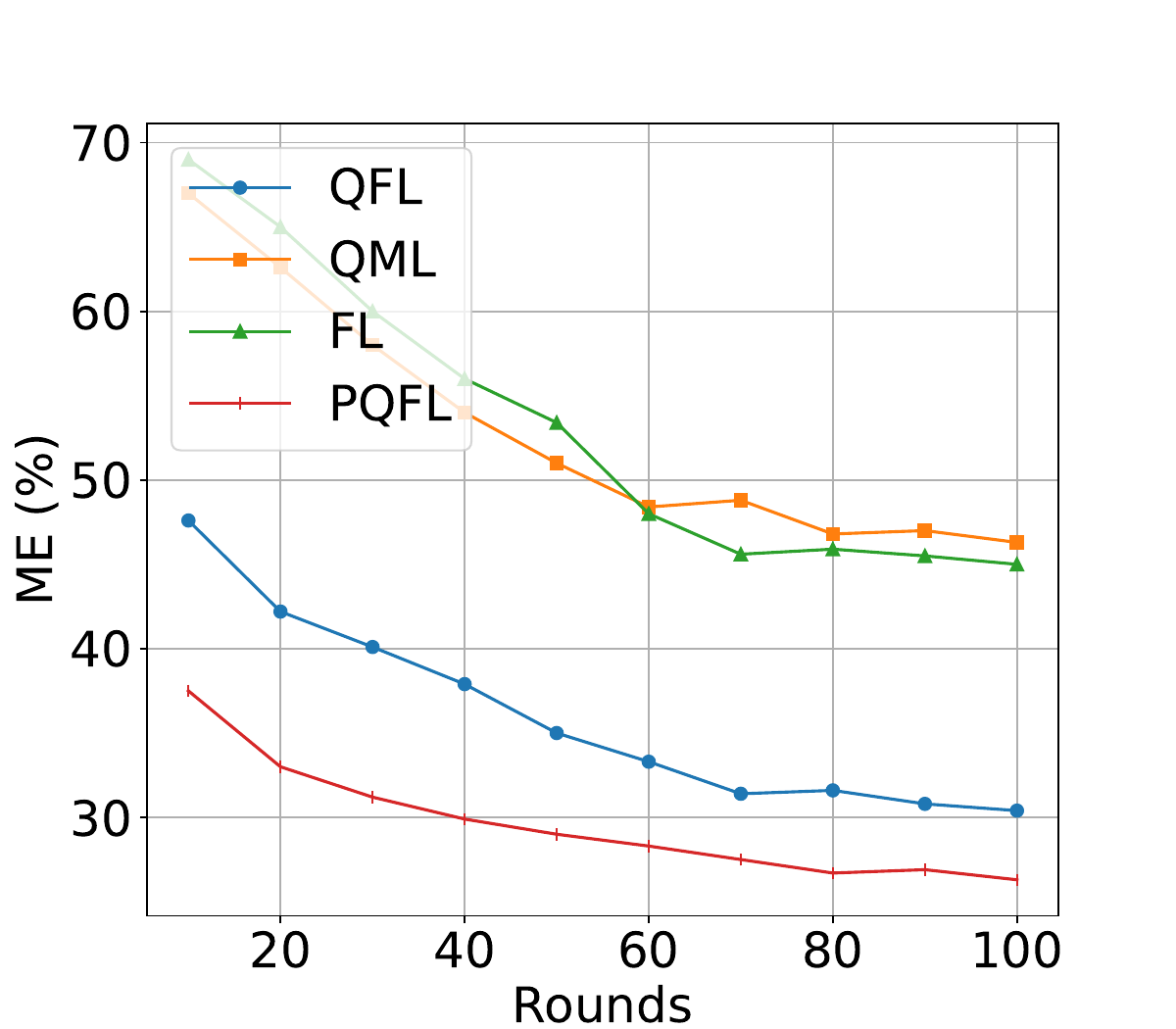}
        \caption{Missing Error}
        \label{fig:nplot2}
    \end{subfigure}%
    \hfill
    \begin{subfigure}{0.247\textwidth}
        \includegraphics[width=\linewidth]{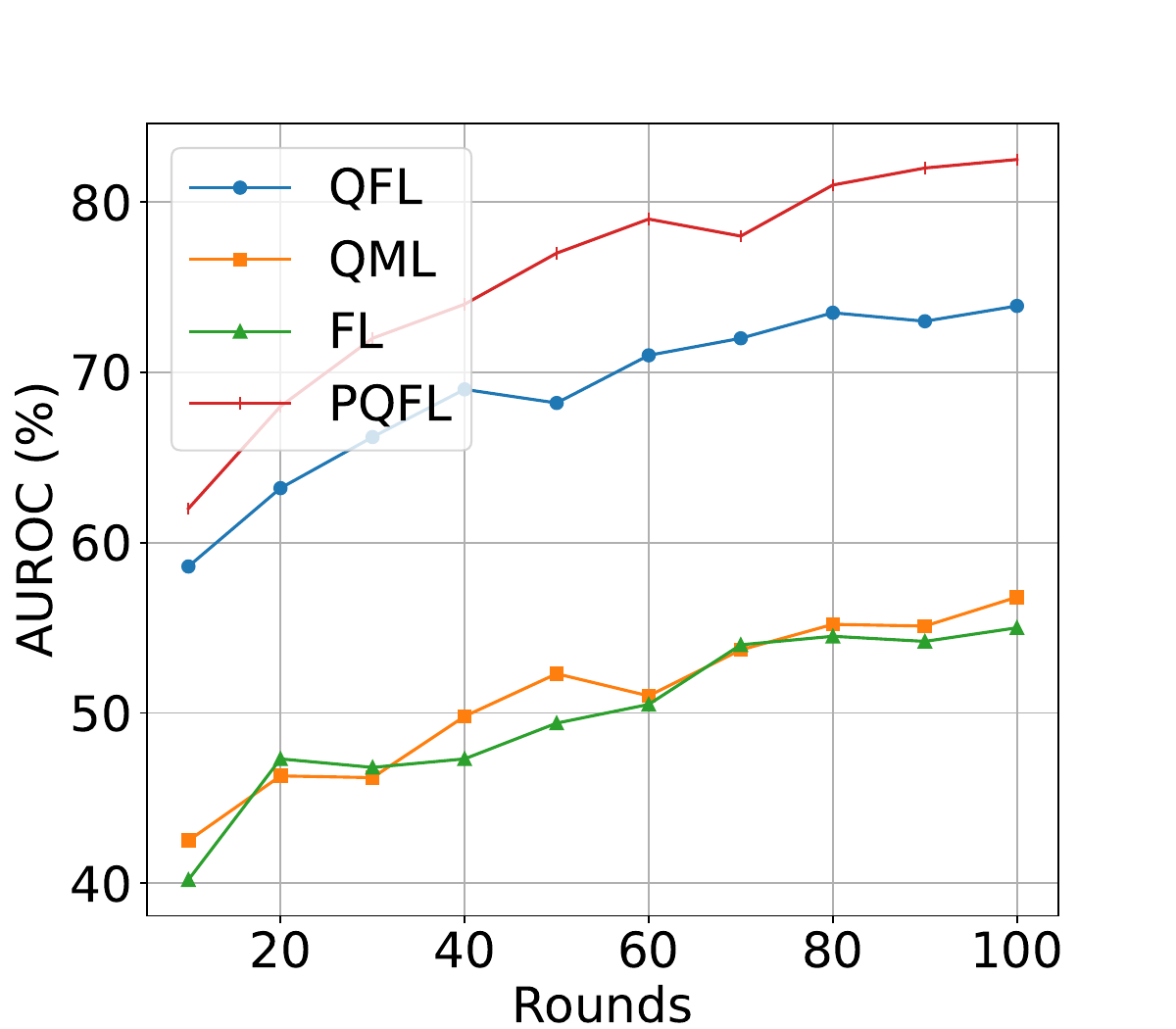}
        \caption{AUROC Score}
        \label{fig:nplot3}
    \end{subfigure}%
    \hfill
    \begin{subfigure}{0.247\textwidth}
        \includegraphics[width=\linewidth]{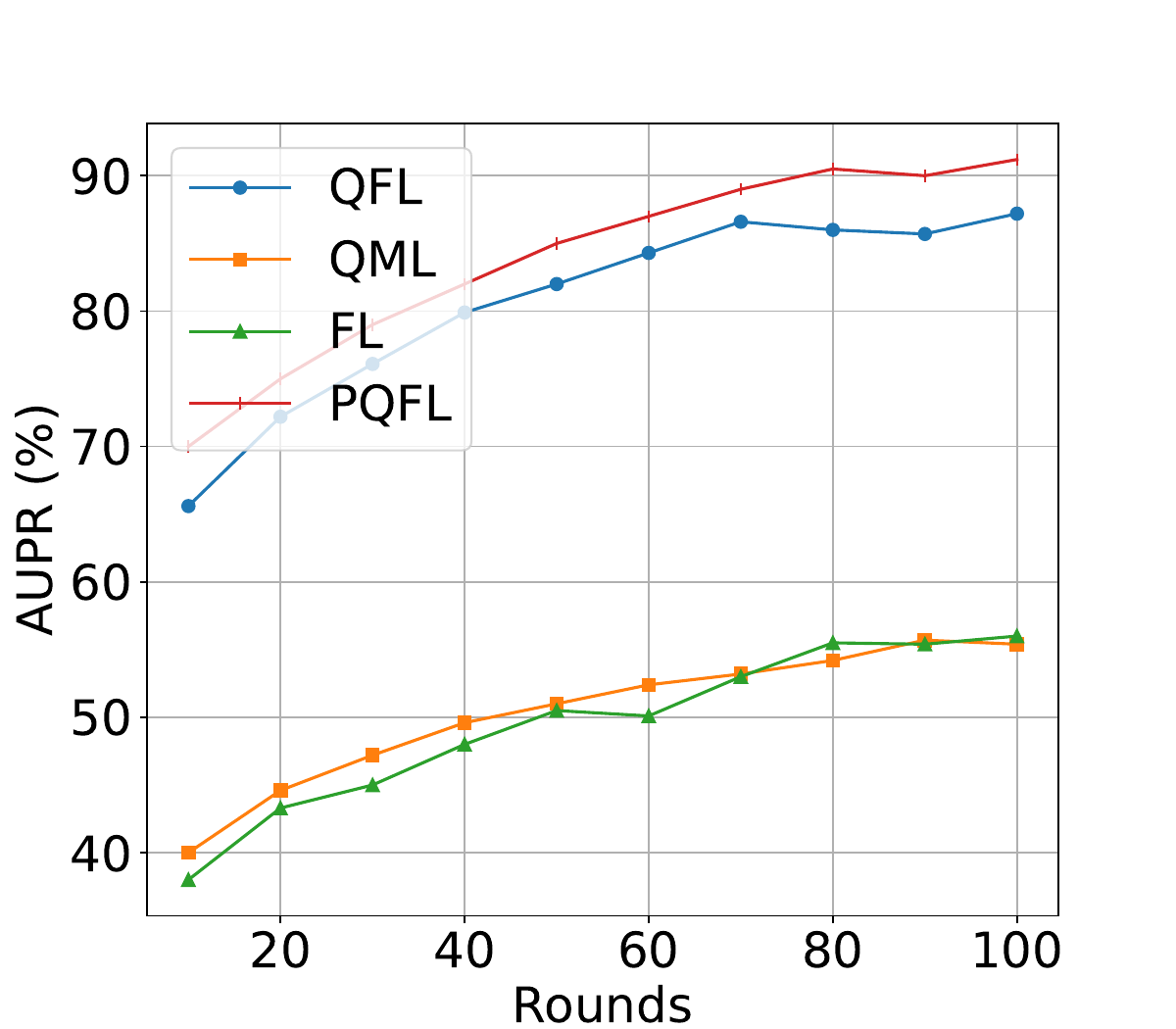}
        \caption{AUPR Score}
        \label{fig:nplot4}
    \end{subfigure}
    \caption{Comparison between different approaches for Anomaly detection, where CIFAR-10 data is used as normal data and CIFAR-100 as anomaly data with 10 clients \textbf{Non-IID data distribution} over 100 global rounds. We compared using false error (FE) in (a), missing error (ME) in (b), AUROC score in (c), and AUPR score in (d). Note that lower FE and ME values and higher AUROC and AUPR scores indicate better performance.}
    \label{fig:niidplots}
\end{figure*}

\subsection{Effects on Using Quantum Model}
To show the effectiveness of the quantum approach, first we plot and compare the performance of QFL, QNN, and FL approaches in terms of FE and ME values in Fig. \ref{fig1}. We use the convolutional neural network (CNN) model for FL and the QNN model for QFL at local clients for 100 global rounds. For the normal dataset, we employ CIFAR-10 data and CIFAR-100 for anomaly data. With the lowest loss values throughout all training rounds, QFL continuously beats QNN and FL in both the FE and ME configurations.  QNN's loss drops from 18.5 to 11.0 under FE and from 35.0 to 19.4 under ME, whereas QFL achieves even lower losses, falling from 15.6 to 7.77 in FE and from 24.6 to 15.2 in ME, even though both approaches show a progressive decline in loss over time.  Furthermore, QFL converges faster than QNN and FL. The simulation results clearly show the superior performance of QFL over FL, as the identical data distribution enables more accurate learning with better convergence speeds for the QNN model. Therefore, we can conclude that a quantum-based model can perform significantly better for anomaly detection than the traditional model.

\begin{table*}[!ht]
  \centering
      \caption{Different number of clients $N$ on anomaly detection using QFL environment across various non-IID data distributions. We have used 3, 5, 10, and 20 clients for comparison, with CIFAR-10 as the normal dataset and CIFAR-100 as the anomaly dataset. Adding more clients often results in increased resilience; therefore, experimenting with the number of clients helps assess scalability, generalization, and convergence.}
  \label{tab:data_distribution_niid}
  \begin{tabular}{|c|c|cc|cc|}
    \toprule
    Data Type & Clients & FE(\%) $\downarrow$ & ME(\%) $\downarrow$ &AUROC(\%) $\uparrow$ & AUPR(\%) $\uparrow$\\
    \midrule
    \multirow{3}{*}{Step} 
    & $N$ = 3 & $14.80 \pm \text{\footnotesize 1.4}$ & $38.22 \pm \text{\footnotesize 1.9}$ & $76.00 \pm \text{\footnotesize 1.0}$ & $87.15 \pm \text{\footnotesize 2.1}$ \\
    & $N$ = 5 & $15.01 \pm \text{\footnotesize 1.2}$ & $37.05 \pm \text{\footnotesize 2.0}$ & $76.24 \pm \text{\footnotesize 0.6}$ & $87.75 \pm \text{\footnotesize 1.7}$ \\
    & $N$ = 10 & $15.30 \pm \text{\footnotesize 1.0}$ & $35.55 \pm \text{\footnotesize 1.1}$ & $76.60 \pm \text{\footnotesize 0.8}$ & $88.05 \pm \text{\footnotesize 1.0}$ \\
    & $N$ = 20 & $15.29 \pm \text{\footnotesize 0.9}$ & $35.49 \pm \text{\footnotesize 1.0}$ & $76.65 \pm \text{\footnotesize 0.9}$ & $88.08 \pm \text{\footnotesize 0.9}$ \\
    \midrule
    \multirow{3}{*}{Dirichlet 0.1} 
    & $N$ = 3 & $10.34 \pm \text{\footnotesize 0.9}$ & $30.26 \pm \text{\footnotesize 1.5}$ & $82.10 \pm \text{\footnotesize 1.1}$ & $90.66 \pm \text{\footnotesize 1.1}$ \\
    & $N$ = 5 & $10.11 \pm \text{\footnotesize 1.1}$ & $30.01 \pm \text{\footnotesize 1.9}$ & $82.25 \pm \text{\footnotesize 1.4}$ & $90.80 \pm \text{\footnotesize 1.6}$ \\
    & $N$ = 10 & $09.89 \pm \text{\footnotesize 1.1}$ & $29.78 \pm \text{\footnotesize 2.4}$ & $82.46 \pm \text{\footnotesize 2.2}$ & $91.04 \pm \text{\footnotesize 0.8}$ \\
    & $N$ = 10 & $09.81 \pm \text{\footnotesize 1.2}$ & $29.77 \pm \text{\footnotesize 0.5}$ & $82.40 \pm \text{\footnotesize 2.1}$ & $91.14 \pm \text{\footnotesize 0.5}$ \\
    \midrule
    \multirow{3}{*}{Dirichlet 0.01} 
    & $N$ = 3 & $07.77 \pm \text{\footnotesize 1.8}$ & $28.22 \pm \text{\footnotesize 1.0}$ & $84.13 \pm \text{\footnotesize 1.9}$ & $92.06 \pm \text{\footnotesize 2.2}$ \\
    & $N$ = 5 & $07.62 \pm \text{\footnotesize 1.4}$ & $27.05 \pm \text{\footnotesize 0.9}$ & $84.29 \pm \text{\footnotesize 1.3}$ & $92.20 \pm \text{\footnotesize 1.3}$ \\
    & $N$ = 10 & $07.59 \pm \text{\footnotesize 0.9}$ & $25.55 \pm \text{\footnotesize 0.7}$ & $84.86 \pm \text{\footnotesize 1.1}$ & $92.44 \pm \text{\footnotesize 1.4}$ \\
    & $N$ = 20 & $07.51 \pm \text{\footnotesize 0.8}$ & $25.48 \pm \text{\footnotesize 0.6}$ & $84.92 \pm \text{\footnotesize 1.0}$ & $92.94 \pm \text{\footnotesize 1.1}$ \\
    \bottomrule
  \end{tabular}
\end{table*}

\subsection{Simulation Results on QFL}
First, we conduct simulation results on IID data in a QFL environment. For the normal dataset, we use CIFAR-10 data, and for the anomaly dataset, we use CIFAR-100 data.  
\begin{table*}[!ht]
\color{black}
    \caption{Different data size $|D|$ on anomaly detection using QFL environment across various non-IID data distributions. The data size includes 10\%, 25\%, 50\%, and 100\%. We have used CIFAR-10 as the normal dataset and CIFAR-100 as the anomaly dataset. }
  \label{tab:data_size}
  \centering
  \begin{tabular}{|c|c|cc|cc|}
    \toprule
    Data Type & Data Size & FE(\%) $\downarrow$ & ME(\%) $\downarrow$ &AUROC(\%) $\uparrow$ & AUPR(\%) $\uparrow$\\
    \midrule
    \multirow{3}{*}{Step} 
    & $|D|$ = 10\% & $32.80 \pm \text{\footnotesize 1.4}$ & $55.22 \pm \text{\footnotesize 1.9}$ & $62.00 \pm \text{\footnotesize 1.0}$ & $74.15 \pm \text{\footnotesize 2.1}$ \\
    & $|D|$ = 25\% & $26.01 \pm \text{\footnotesize 1.2}$ & $45.05 \pm \text{\footnotesize 2.0}$ & $67.24 \pm \text{\footnotesize 0.6}$ & $79.75 \pm \text{\footnotesize 1.7}$ \\
    & $|D|$ = 50\% & $19.07 \pm \text{\footnotesize 1.1}$ & $40.40 \pm \text{\footnotesize 1.4}$ & $73.44 \pm \text{\footnotesize 0.9}$ & $84.50 \pm \text{\footnotesize 1.2}$ \\
    & $|D|$ = 100\% & $15.30 \pm \text{\footnotesize 1.0}$ & $35.55 \pm \text{\footnotesize 1.1}$ & $76.60 \pm \text{\footnotesize 0.8}$ & $88.05 \pm \text{\footnotesize 1.0}$ \\
    \midrule
    \multirow{3}{*}{Dirichlet 0.1} 
    & $|D|$ = 10\% & $23.32 \pm \text{\footnotesize 1.9}$ & $52.26 \pm \text{\footnotesize 1.5}$ & $64.10 \pm \text{\footnotesize 1.1}$ & $80.23 \pm \text{\footnotesize 2.1}$ \\
    & $|D|$ = 25\% & $16.11 \pm \text{\footnotesize 2.1}$ & $40.01 \pm \text{\footnotesize 1.9}$ & $73.25 \pm \text{\footnotesize 1.4}$ & $86.66 \pm \text{\footnotesize 1.6}$ \\
    & $|D|$ = 50\% & $11.22 \pm \text{\footnotesize 1.1}$ & $34.10 \pm \text{\footnotesize 2.0}$ & $79.64 \pm \text{\footnotesize 0.9}$ & $89.80 \pm \text{\footnotesize 1.8}$ \\
    & $|D|$ = 100\% & $09.89 \pm \text{\footnotesize 1.0}$ & $29.78 \pm \text{\footnotesize 1.4}$ & $82.46 \pm \text{\footnotesize0.8}$ & $91.04 \pm \text{\footnotesize 0.8}$ \\
    \midrule
    \multirow{3}{*}{Dirichlet 0.01} 
    & $|D|$ = 10\% & $22.55 \pm \text{\footnotesize 1.2}$ & $51.22 \pm \text{\footnotesize 1.7}$ & $66.13 \pm \text{\footnotesize 1.9}$ & $86.06 \pm \text{\footnotesize 2.1}$ \\
    & $|D|$ = 25\% & $14.83 \pm \text{\footnotesize 1.6}$ & $38.05 \pm \text{\footnotesize 1.3}$ & $74.29 \pm \text{\footnotesize 1.3}$ & $89.20 \pm \text{\footnotesize 2.3}$ \\
    & $|D|$ = 50\% & $08.48 \pm \text{\footnotesize 0.8}$ & $27.37 \pm \text{\footnotesize 1.2}$ & $80.10 \pm \text{\footnotesize 1.0}$ & $91.72 \pm \text{\footnotesize 1.0}$ \\
    & $|D|$ = 100\% & $07.59 \pm \text{\footnotesize 0.7}$ & $25.55 \pm \text{\footnotesize 0.9}$ & $84.86 \pm \text{\footnotesize 1.1}$ & $92.44 \pm \text{\footnotesize 1.2}$ \\
    \bottomrule
  \end{tabular}

\end{table*}

\begin{table*}[t]
\color{black}
\centering
\caption{Ablation Studies on Personalization Strategies in PQFL. 
Each variant removes or alters one personalization component to isolate its contribution. 
Metrics: Final Error (FE, $\downarrow$), Model Error (ME, $\downarrow$), AUROC ($\uparrow$), and AUPR ($\uparrow$).}
\label{tab:ablation_personalization}
\renewcommand{\arraystretch}{1.2}
\setlength{\tabcolsep}{5.5pt}
\begin{tabular}{|l|c|c|c|c|}
\hline
\textbf{Variant} & \textbf{FE ($\downarrow$)} & \textbf{ME ($\downarrow$)} & \textbf{AUROC ($\uparrow$)} & \textbf{AUPR ($\uparrow$)} \\ \hline
No Personalization ($\lambda=0$)          & 18.6 $\pm$ 1.1 & 21.4 $\pm$ 1.2 & 76.2 $\pm$ 0.8 & 82.4 $\pm$ 0.7 \\ 
Classical FedProx Regularization          & 15.9 $\pm$ 1.0 & 18.2 $\pm$ 1.1 & 80.8 $\pm$ 0.6 & 86.2 $\pm$ 0.5 \\ 
Uniform Encoding (no heterogeneity)       & 14.2 $\pm$ 0.9 & 16.8 $\pm$ 1.0 & 82.5 $\pm$ 0.5 & 87.1 $\pm$ 0.4 \\ 
\textbf{Proposed PQFL (full)}             & \textbf{12.8 $\pm$ 0.8} & \textbf{15.0 $\pm$ 1.0} & \textbf{85.4 $\pm$ 0.5} & \textbf{90.5 $\pm$ 0.4} \\ \hline
\end{tabular}
\end{table*}

\begin{table}[t]
\small
\centering
\color{black}
\caption{Comparison of communication and computational resource cost between QFL and PQFL ($N_c{=}10$, $K{=}50$, $D_q{=}4$, $\ell{=}3$, $b{=}32$ bits).}
\label{tab:comm_resource_compare}
\renewcommand{\arraystretch}{1.15}
\setlength{\tabcolsep}{4pt}
\begin{tabular}{|l|c|c|c|}
\hline
\textbf{Metric} & \textbf{QFL} & \textbf{PQFL} \\
\hline
Model parameters per client ($D$) & $12$ & $12$ \\
Per-round payload (bits) & $2Db = 768$ & $2Db + 0.1D$   \\
Computation time / round (ms) & 9.8 & 10.3  \\
Energy / round (µJ, Wi-Fi) & 0.75 & 0.79 \\
Convergence rounds (to target loss) & 70 & 55  \\
Total energy to converge (µJ) & 52.5 & 43.5  \\
\hline
\end{tabular}
\end{table}

\begin{table*}[ht]
    \centering
        \caption{Comparison results between different quantum noise levels $\epsilon$ in \textit{PQFL} and QFL anomaly detection for 10 clients' non-IID data distribution. We have used CIFAR-10 as the normal dataset and CIFAR-100 as the anomaly dataset. The noise levels used in this comparison are 0.5, 0.1, 0.01, and 0.001, with 0.5 adding the most noise and 0.001 the least. }
    \label{tab: noise}
    \begin{tabular}{|p{0.70cm}|p{1.4cm}|p{1.4cm}|p{2cm}|p{1.8cm}|p{1.4cm}|p{1.4cm}|p{2cm}|p{1.8cm}|}
    \hline
    $\multirow{2}{*}{$\epsilon$}$ & \multicolumn{4}{c|}{\textit{PQFL}} &\multicolumn{4}{c|}{QFL} \\
        \cline{2-9}
         & FE(\%) $\downarrow$ & ME(\%) $\downarrow$ & AUROC(\%) $\uparrow$ & AUPR(\%) $\uparrow$ & FE(\%) $\downarrow$ & ME(\%) $\downarrow$ & AUROC(\%) $\uparrow$ & AUPR(\%) $\uparrow$ \\
        \hline
        0.5  & 24.1$\pm$1.3  & 42.6$\pm$0.7  & 62.2$\pm$2.6 & 65.5$\pm$1.4 & 28.5$\pm$2.3  & 45.6$\pm$2.7  & 61.2$\pm$1.6 & 65.2$\pm$3.6  \\
        0.1  & 18.4$\pm$1.8  & 35.5$\pm$1.5  & 70.8$\pm$1.9  & 72.0$\pm$2.0 & 21.1$\pm$1.8  & 38.9$\pm$1.4  & 70.1$\pm$2.9  & 71.5$\pm$2.1  \\
        0.01  & 10.4$\pm$2.6  & 29.4$\pm$1.7  & 78.5$\pm$0.9  & 80.8$\pm$1.9 & 13.5$\pm$1.0  & 31.0$\pm$2.7  & 78.5$\pm$1.2  & 78.8$\pm$1.2 \\
        0.001 &  \textbf{8.29$\pm$1.4} & \textbf{25.72$\pm$0.6} & \textbf{84.2$\pm$0.9} & \textbf{91.88$\pm$0.9} &  \textbf{9.01$\pm$1.1} & \textbf{29.7$\pm$0.9} & \textbf{81.1$\pm$1.0} & \textbf{88.66$\pm$0.7} \\ 
        \hline
    \end{tabular}
\end{table*}
\textbf{Quantum hyperparameter analysis.}
We conduct ablation experiments across different quantum parameters in our QFL model. For hyperparameters, we have emphasized parameters that effectively affect the performance of the quantum model, such as quantum bits (qubits) and quantum layers. First, we compare the performance using different qubits $D_q$ in each communication round. We present our findings in Table \ref{tab: qubits}. All settings have the same parameters and training epochs, other than qubits. For comparison, we set the value of $D_q$ to 2,3,5, and 10. The result shows that $D_q = 10$ provided the best performance, and with more qubits, the QFL model yields a better performance. Subsequent simulation results are conducted using 10 qubits.  


We also evaluate the impact of quantum layers $l$ in our QFL model. For comparison, we set the value of $l$ to 1,2,3,5, and 10. The simulation result is shown in Table~\ref{tab: qlayer}. For this experiment, all values are set to be constant, including training rounds and $D_c = 10$. The result demonstrates that we can not draw any conclusion about qubits for quantum layers, as the performance difference is not linear. However, $l=1$ has performed the best, and we set the value of $l$ to 1 for our next experiments.

\subsection{Simulation Results on \textit{PQFL}}
\begin{table*}[t]
  \centering
  \caption{Sensitivity of PQFL to the regularization parameter $\lambda$ (mean$\pm$std over 5 seeds). Lower is better for FE/ME; higher is better for AUROC/AUPR. Best values in \textbf{bold}.}
  \label{tab:lambda_sweep}
  \setlength{\tabcolsep}{5pt}
  \begin{tabular}{lcccccccc}
    \toprule
    $\lambda \to$ & 0 & $1\mathrm{e}{-4}$ & $1\mathrm{e}{-3}$ & $3\mathrm{e}{-3}$ & $1\mathrm{e}{-2}$ & $3\mathrm{e}{-2}$ & \textbf{0.1} & 0.3 \\
    \midrule
    AUROC (\%) $\uparrow$ &
    $72.10 \pm \text{\footnotesize 0.9}$ &
    $72.80 \pm \text{\footnotesize 0.9}$ &
    $74.50 \pm \text{\footnotesize 0.9}$ &
    $75.10 \pm \text{\footnotesize 0.8}$ &
    $75.90 \pm \text{\footnotesize 0.8}$ &
    $76.20 \pm \text{\footnotesize 0.8}$ &
    \textbf{76.60 $\pm \text{\footnotesize 0.8}$} &
    $76.00 \pm \text{\footnotesize 0.8}$ \\
    AUPR (\%) $\uparrow$ &
    $82.40 \pm \text{\footnotesize 1.2}$ &
    $83.10 \pm \text{\footnotesize 1.2}$ &
    $85.20 \pm \text{\footnotesize 1.1}$ &
    $86.10 \pm \text{\footnotesize 1.0}$ &
    $86.90 \pm \text{\footnotesize 1.0}$ &
    $87.60 \pm \text{\footnotesize 1.0}$ &
    \textbf{88.05 $\pm \text{\footnotesize 1.0}$} &
    $87.30 \pm \text{\footnotesize 1.0}$ \\
    FE (\%) $\downarrow$ &
    $19.80 \pm \text{\footnotesize 1.3}$ &
    $19.20 \pm \text{\footnotesize 1.2}$ &
    $17.90 \pm \text{\footnotesize 1.1}$ &
    $17.10 \pm \text{\footnotesize 1.1}$ &
    $16.40 \pm \text{\footnotesize 1.0}$ &
    $15.80 \pm \text{\footnotesize 1.0}$ &
    \textbf{15.30 $\pm \text{\footnotesize 1.0}$} &
    $16.00 \pm \text{\footnotesize 1.1}$ \\
    ME (\%) $\downarrow$ &
    $41.20 \pm \text{\footnotesize 1.6}$ &
    $40.80 \pm \text{\footnotesize 1.5}$ &
    $39.40 \pm \text{\footnotesize 1.3}$ &
    $38.80 \pm \text{\footnotesize 1.3}$ &
    $37.60 \pm \text{\footnotesize 1.2}$ &
    $36.40 \pm \text{\footnotesize 1.1}$ &
    \textbf{35.55 $\pm \text{\footnotesize 1.1}$} &
    $36.80 \pm \text{\footnotesize 1.2}$ \\
    \bottomrule
  \end{tabular}
\end{table*}

The simulation results on QFL clearly demonstrate the enhanced performance of quantum physics over the FL framework in terms of higher accuracy and faster convergence on IID data. However, in the heterogeneous environment, the QFL framework struggles to achieve stable convergence, resulting in poor globalization of the global model. To further improve the results, especially for the non-IID data distribution, we have introduced \textit{PQFL}. 
To overcome this limitation, we provide a regularization-based approach for \textit{PQFL}, which penalizes excessive variation between local and global models.  The method performs better on non-IID datasets by finding a compromise between utilizing global insights and maintaining local adaptability, guaranteeing strong convergence and better overall anomaly detection outcomes. PQFL has a U-shaped response, with tiny $\lambda$ over-personalizing and high $\lambda$ over-regularizing. A broad plateau develops for $\lambda\in [0.03,0.2]$. At $\lambda{=}0.1$, the ideal trade-off is with FE $15.30\pm\text{\footnotesize1.0}\%$, ME $35.55\pm\text{\footnotesize1.1}\%$, AUROC $76.60\pm\text{\footnotesize0.8}\%$, and AUPR $88.05\pm\text{\footnotesize1.0}\%$; the same pattern holds.

\textbf{Comparison between different approaches.}
The simulation results compare four distinct learning algorithms (QFL, QML, FL, and \textit{PQFL}) across numerous rounds using four essential metrics: false error (FE), missing error (ME), AUROC, and AUPR in Fig.~\ref{fig:niidplots}.  \textit{PQFL} has greater resilience and efficiency, consistently outperforming other approaches across all measurements.  \textit{PQFL} has the lowest FE and ME values in terms of error reduction with increasing rounds, which suggests that it is more accurate in identifying missing and erroneous errors.  Interestingly, \textit{PQFL} has a FE of 8.2\% at 100 rounds, which is much lower than FL (31\%), QML (15\%), and QFL (10.2\%).  Similarly, in ME, \textit{PQFL} continues to have the lowest error at 26.3\%, exceeding FL (45\%) and QML (46.3\%), but QFL stays marginally better at 30.4\%. \textit{PQFL} continuously obtains the highest values when AUROC and AUPR are evaluated, demonstrating exceptional classification performance.  At 100 rounds, AUROC rises gradually to 82.5\%, outperforming FL (55\%), QML (56.8\%), and QFL (73.9\%).  Comparably, \textit{PQFL}'s AUPR of 91.2\% shows remarkable precision and recall, but QFL, QML, and FL fall short at 87.2\%, 55.4\%, and 56\%, respectively.  These patterns show that \textit{PQFL} is the most successful of the four methods since it offers faster convergence, fewer errors, and superior classification performance.

\textbf{Effect on number of clients.}
Table \ref{tab:data_distribution_niid} compares the impact of varying client numbers ($N$) in a QFL environment on anomaly detection using CIFAR-10 (normal) and CIFAR-100 (anomaly) datasets. Increasing N from 3 to 10 increases overall performance across all distributions. In the non-IID distribution, FE drops from 14.80\% ($N=3$) to 15.30\% ($N=10$), ME falls from 38.22\% to 35.55\%, while AUROC marginally improves from 76.00\% to 76.60\%. These findings show that increasing the number of clients improves anomaly detection accuracy, especially in strongly non-IID distributions, hence enhancing both false error reduction and classification performance.

\textcolor{black}{\textbf{Analyzing stress testing.} 
Table \ref{tab:data_size} compares the effect of different data sizes $|D|$ on anomaly detection performance in a QFL environment, utilizing CIFAR-10 (normal) and CIFAR-100 (anomaly) datasets from a non-IID distribution. Increasing the data amount from 10\% to 100\% greatly increases performance by lowering False Errors (FE) and Missing Errors (ME) while enhancing AUROC and AUPR. In the non-IID distribution, FE lowers from 32.80\% (10\%) to 15.30\% (100\%), ME decreases from 55.22\% to 35.55\%, while AUROC increases from 62.00\% to 76.60\%. AUPR shows a similar pattern, improving by 14-18 percentage points across distributions. These findings show that larger datasets greatly improve anomaly detection accuracy, especially in highly non-IID environments, by lowering errors and enhancing generalization.}

\begin{table}
    \centering
        \caption{Comparison on different measurements $M$ in \textit{PQFL} to tackle the shot noise in quantum. For comparison, we use measurement values of 1,50, and 100. Fewer quantum measurements result in suboptimal performance, whereas more measurements might improve representation, but create significant complexity and require quantum resources. We use 10 clients' non-iid CIFAR-10 normal data and CIFAR-100 anomaly data. }
    \label{tab: measurement}
    \begin{tabular}{p{0.30cm}|p{1.30cm}|p{1.40cm}|p{2.0cm}|p{1.80cm}}
        \hline
        $M$ & FE(\%) $\downarrow$ & ME(\%) $\downarrow$ & AUROC(\%) $\uparrow$ & AUPR(\%) $\uparrow$ \\
        \hline
        1  & 8.29$\pm$1.4 & 25.72$\pm$0.6 & 84.2$\pm$0.9 & 91.88$\pm$2.9  \\
        50  & 8.37$\pm$1.2  & 25.2$\pm$0.7  & 85.2$\pm$0.6 & 92.5$\pm$1.6  \\
        100  & 7.60$\pm$0.7  & 24.9$\pm$0.2  & 86.3$\pm$0.3  & 93.0$\pm$0.4  \\
        \hline
    \end{tabular}
\end{table}
\begin{table*}[!ht]
\color{black}
\centering
\caption{Result comparison of Anomaly detection between different state-of-the-art approaches. For comparison, we selected CNN\cite{yan2023fgdae, wang2024real}, QML \cite{wang2022data,kukliansky2024network}, FL\cite{wang2024regression, dong2024fadngs}, QFL, and our proposed \textit{PQFL} methods in \textbf{non-IID data distribution}. We have used CIFAR-10 as normal data and CIFAR-100, SVHN, and ImageNet as anomaly datasets, and have applied all the approaches in each combination.}
\label{Table: stateoftheart_niid}
\footnotesize
\begin{tabular}{@{}|p{1.4cm}|p{1.6cm}|c|c|c|c|c|@{}}
\toprule
Normal Dataset & Anomaly Dataset & Method & FE (\%) $\downarrow$ & ME (\%) $\downarrow$ &AUROC (\%) $\uparrow$ & AUPR (\%) $\uparrow$ \\
\midrule
\multirow{15}{*}{CIFAR-10} & \multirow{5}{*}{CIFAR-100} & CNN \cite{yan2023fgdae, wang2024real} & 49.6 $\pm$ 0.8 & 53.9 $\pm$ 1.2 & 55.8 $\pm$ 0.9 & 52.0 $\pm$ 2.0 \\
                                 && QML \cite{wang2022data,kukliansky2024network, zhu2023anomaly} & 46.8 $\pm$ 1.2 & 52.9 $\pm$ 0.9 & 56.4 $\pm$ 0.9                        & 53.9 $\pm$ 0.9                      \\
                                &      & FL\cite{wang2024regression, dong2024fadngs}     & 31.2 $\pm$ 1.8                       & 45.3 $\pm$ 1.9                       & 56.8 $\pm$ 1.2                        & 55.2 $\pm$ 1.3                      \\
                                 &         & QFL & 10.2 $\pm$ 1.2                       & 30.4 $\pm$ 0.8                       & 73.9 $\pm$ 1.5                        & 87.2 $\pm$ 1.2                      \\
                                 &         & \textit{PQFL}     & \textbf{8.2 $\pm$ 0.7}                       & \textbf{26.3 $\pm$ 0.3}                       & \textbf{82.5 $\pm$ 0.8}                        & \textbf{91.2 $\pm$ 0.4}                      \\
                                 \cline{2-7}
 & \multirow{5}{*}{SVHN}      & CNN \cite{yan2023fgdae, wang2024real}  & 40.6 $\pm$ 2.8 & 29.8 $\pm$ 1.9 & 55.8 $\pm$ 0.9 & 56.0 $\pm$ 2.0 \\
                                 && QML \cite{wang2022data,kukliansky2024network, zhu2023anomaly} & 38.2 $\pm$ 1.3 & 26.9 $\pm$ 1.5 & 92.9 $\pm$ 1.9                        & 91.9 $\pm$ 1.8                      \\
                                &      & FL\cite{wang2024regression, dong2024fadngs}    & 12.9 $\pm$ 1.7                       & 19.3 $\pm$ 1.7                       & 93.2 $\pm$ 1.4                        & 91.2 $\pm$ 1.5                      \\
                                 &         & QFL & 12.6 $\pm$ 1.2                       & 12.4 $\pm$ 0.9                       & 95.9 $\pm$ 1.0                        & 95.2 $\pm$ 1.0                      \\
                                 &         & \textit{PQFL} & \textbf{8.6 $\pm$ 0.6}& \textbf{7.3 $\pm$ 0.7}   & \textbf{97.5 $\pm$ 0.8} & \textbf{96.2 $\pm$ 0.5}                      \\
                                 \cline{2-7}
 & \multirow{5}{*}{\shortstack{ImageNet}}  & CNN\cite{yan2023fgdae, wang2024real}  & 40.9 $\pm$ 1.7 & 45.1 $\pm$ 1.1 & 55.8 $\pm$ 0.9 & 57.1 $\pm$ 2.2 \\
                                 && QML \cite{wang2022data,kukliansky2024network, zhu2023anomaly} & 41.2 $\pm$ 1.4 & 44.4 $\pm$ 0.9 & 54.4 $\pm$ 1.9                        & 55.9 $\pm$ 2.9                      \\
                                &      & FL\cite{wang2024regression, dong2024fadngs}   & 26.8 $\pm$ 1.0                       & 22.2 $\pm$ 1.2                       & 54.8 $\pm$ 1.2                        & 77.2 $\pm$ 1.9                      \\
                                 &         & QFL & 13.9 $\pm$ 1.6                       & 17.1 $\pm$ 0.8                       & 82.2 $\pm$ 1.2                        & 86.8 $\pm$ 1.2                      \\
                                 &         & \textit{PQFL}     & \textbf{12.8 $\pm$ 0.8}                       & \textbf{15.0 $\pm$ 1.0}                       & \textbf{85.4 $\pm$ 0.5}                        & \textbf{90.5 $\pm$ 0.4 }                     \\
                                 \bottomrule
\end{tabular}
\end{table*}
\textcolor{black}{
\textbf{Ablation on personalization strategies.} We conduct an ablation study comparing three simplified variants of PQFL in order to quantify the contribution of each personalization element in Eq.~(14): (i) removing the proximal term ($\lambda=0$), (ii) substituting a classical FedProx-style regularizer for the quantum-aware regularization, and (iii) enforcing identical quantum encodings across all clients (no encoding heterogeneity).
 Table~\ref{tab:ablation_personalization} demonstrates that eliminating or simplifying the personalization component consistently reduces performance across all measures.  The entire PQFL model achieves the lowest Final Error (FE) and Model Error (ME), as well as the greatest AUROC and AUPR, demonstrating the importance of circuit-aware personalization and encoding-level adaptation for steady convergence and robust anomaly detection in non-IID environments.}

\textcolor{black}{\textbf{Communication cost.} The communication and energy profiles of QFL and PQFL are compared in Table \ref{tab:comm_resource_compare} with the identical 10-client, 50-round configuration.  Transmitting the regularization coefficient only slightly raises PQFL's communication cost ($\approx$1.6\%) since the two frameworks share the same parameter tensors.  Although PQFL converges in 21\% fewer rounds, the local L2 update adds around 5\% more computation and energy every round, which results in a net 17\% reduction in overall energy usage.  As a result, PQFL achieves more efficiency per global convergence, exchanging a little more work every round for significantly quicker global alignment and lower overall resource consumption.}

\subsection{Comparison with Quantum Noise}
\textbf{Effect on noisy data.}
To assess the robustness of PQFL and QFL under realistic noise, we simulate the algorithms using \texttt{Qiskit-FGDAER} with a noise model built from IBM's superconducting quantum backend (e.g., \texttt{ibmq_guadalupe}) and execute them on actual hardware.  The backend reports average single-qubit gate error rates of $3.5 \times 10^{-3}$, two-qubit gate (CNOT) error rates of $1.2 \times 10^{-2}$, decoherence times $T_1 \approx 100 \, \mu$s and $T_2 \approx 80 \, \mu$s, and crosstalk probabilities of $5 \times 10^{-3}$ for adjacent qubit pairs, justifying the noise model's fidelity to real hardware.
Table \ref{tab: noise} compares the effect of varying quantum noise levels ($\epsilon$) using \texttt{Qiskit-FGDAER} noise model on anomaly detection performance in \textit{PQFL} and QFL using CIFAR-10 (normal) and CIFAR-100 (anomaly) datasets for 10 clients in a non-IID setting. As noise falls from $\epsilon=0.5$ to $\epsilon=0.001$, performance increases dramatically. In \textit{PQFL}, False Error (FE) falls from 24.1\% (0.5) to 8.29\% (0.001), Missing Error (ME) decreases from 42.6\% to 25.72\%, AUROC improves from 62.2\% to 84.2\%, and AUPR increases from 65.5\% to 91.88\%. Similarly, in QFL, FE falls from 28.5\% to 9.01\%, ME from 45.6\% to 29.7\%, AUROC from 61.2\% to 81.1\%, and AUPR from 65.2\% to 88.66\%. \textit{PQFL} consistently beats QFL across all noise levels, demonstrating greater robustness to quantum noise. At greater noise levels ($\epsilon=0.5$), \textit{PQFL} outperforms QFL with 4.4\% lower FE and 3.0\% lower ME. At $\epsilon=0.001$, \textit{PQFL} outperforms QFL with an AUROC of 84.2\%, whereas QFL only reaches 81.1\%. These findings show that decreased quantum noise leads to improved anomaly identification and that \textit{PQFL} is more resilient to noise than QFL, making it more successful in noisy quantum environments.

\textbf{Effects on quantum measurements.}
Table \ref{tab: measurement} compares performance in \textit{PQFL} with different quantum measurement counts ($M$).  It shows how quantum measurements might reduce the impacts of quantum shot noise, which is a fundamental type of noise in quantum systems caused by the probabilistic nature of quantum state measurement.  Shot noise can greatly reduce the accuracy of quantum state estimations, resulting in large inaccuracies in quantum computations.

\subsection{Comparison with State-of-the-art Approaches}

We compare the results of our proposed PQFL approach with other state-of-the-art methods, including CNN\cite{yan2023fgdae, wang2024real}, QML \cite{wang2022data,kukliansky2024network}, FL\cite{wang2024regression, dong2024fadngs}, and QFL for non-IID data distribution in Table \ref{Table: stateoftheart_niid}. When CIFAR-10 is used as the normal dataset and CIFAR-100 as the anomalous dataset, \textit{PQFL} has the lowest False Error (FE) of 8.2\% and Missing Error (ME) of 26.3\%, much better than QFL (10.2\% FE, 30.4\% ME) and far surpassing FL (31.2\% FE, 45.3\% ME).  The efficacy of conventional and quantum machine learning methods in this context is shown by CNN and QML's further decreased performance.  Furthermore, \textit{PQFL} outperforms the other approaches in terms of AUROC (82.5\%) and AUPR (91.2\%), demonstrating that it offers a more accurate and dependable anomaly detection mechanism.

\textit{PQFL} continues to outperform rival approaches when the methods are applied to the SVHN (anomaly dataset) and CIFAR-10 (normal dataset).  It outperforms FL (12.9\% FE, 19.3\% ME) and QFL (12.6\% FE, 12.4\% ME), although it has the lowest FE (8.6\%) and ME (7.3\%).  Furthermore, \textit{PQFL} achieves the greatest AUROC (97.5\%) and AUPR (96.2\%), demonstrating its remarkable capacity to differentiate anomalous data from typical data.  The enhancements over FL (93.2\% AUROC, 91.2\% AUPR) and QFL (95.9\% AUROC, 95.2\% AUPR) further confirm its resilience in managing a variety of complex non-IID data distributions.

\textcolor{black}{Similarly, \textit{PQFL} continues to outperform QFL (13.9\% FE, 17.1\% ME) and FL (26.8\% FE, 22.2\% ME) for CIFAR-10 (normal) and ImageNet (anomaly dataset), obtaining FE of 12.8\% and ME of 15.0\%.  Its dependability is further supported by the fact that \textit{PQFL} still differs significantly from traditional CNN/QML-based techniques.  Notably, \textit{PQFL} outperforms QFL (82.2\% AUROC, 86.8\% AUPR) and FL (54.8\% AUROC, 77.2\% AUPR) in terms of capturing anomalies, as seen by its AUROC (85.4\%) and AUPR (90.5\%). }

Therefore, in comparison to CNN, QML, FL, and QFL, \textit{PQFL} consistently produces the lowest error rates and the best classification performance across all three anomaly datasets.  The method's resilience in anomaly detection is demonstrated by its capacity to maximize AUROC and AUPR while minimizing false and missing errors.  This superior performance is particularly noticeable in non-IID circumstances, where conventional methods underperform.  \textit{PQFL} is a promising method for real-world anomaly detection tasks because of its versatility across a variety of datasets, which indicates its efficacy in privacy-preserving federated learning applications.

\section{Limitations}
{\color{black}

\textbf{(1) Quantum hardware availability.}  
As genuine quantum devices have restricted accessibility and queue latency, most current studies rely on simulator backends.  Although calibration data from IBM backends was utilized to simulate realistic noise, large-scale validation on actual hardware is still limited by qubit counts, coherence durations, and circuit depth.

\textbf{(2) Scalability to larger client populations.}  
To ensure computational tractability, we limit our investigation to a reasonable number of customers ($N_c=10$).  To offset the exponential rise in simulation time and circuit assessments, certain PQFL applications may need to be scaled to hundreds of clients by minimizing communication costs, adaptive aggregation, and hybrid classical-quantum resource scheduling.

\textbf{(3) Stability of personalization under noise.}  
The personalization technique $\lambda(w_{n}-w_{g})$ demonstrates consistent convergence under controlled non-IID noise.  However, the stability of personal updates may worsen in the presence of genuine hardware noise, which includes stochastic variations in gate integrity, measurement error, and readout drift.  Future developments should include dynamic noise calibration, error mitigation feedback loops, and noise-adaptive regularization.

\textbf{4) Only considering data heterogeneity for simulation.} \textcolor{black}{Federated optimization is a well-established concept in classical machine learning, but less is known about its quantum equivalents, particularly when dealing with actual noise and heterogeneous encoding.  Instead of using essentially different hardware designs, heterogeneity is represented in this work using non-IID data and varied noise channels across identical PQC architectures.  Therefore, rather than being a comprehensive solution to heterogeneous-hardware aggregation, the suggested framework should be seen as a quantum-influenced adaptation of federated personalization principles.}


\section{Conclusion}
In this work, we addressed the weaknesses of conventional learning, QML, and QFL by proposing (\textit{PQFL}) as a novel method for anomaly identification.  Our approach successfully addresses the difficulties introduced by non-IID data distributions, guaranteeing that the global model can accommodate a variety of client datasets while preserving the advantages of federated learning in terms of security and privacy. Extensive simulations show that \textit{PQFL} performs better than existing methods concerning False Error (FE), Missing Error (ME), AUROC, and AUPR across a variety of dataset combinations, including CNN, QML, FL, and QFL.  Our findings demonstrate that \textit{PQFL} performs better in classification across a range of anomaly detection scenarios, achieving faster convergence and reduced error rates.  The simulation results show that the proposed \textit{PQFL} approach improves by up to 23.4\%, 19\%, 24.2\%, and 20.5\% in terms of false error (FE), missing error (ME), area under the receiver operating characteristic curve (AUROC), and area under the precision-recall curve (AUPR), respectively.

\bibliography{main}
\bibliographystyle{IEEEtran}

\begin{IEEEbiographynophoto}{Ratun Rahman} is a Ph.D. candidate in the Department of Electrical and Computer Engineering at The University of Alabama in Huntsville, USA. His work focuses on machine learning, federated learning, and quantum machine learning. He has published papers in several IEEE journals including IEEE TVT, IEEE IoTJ, and IEEE GRSL, and conferences including NeurIPS and CVPR workshops, IEEE QCE, and IEEE CCNC. 
\end{IEEEbiographynophoto}

\begin{IEEEbiographynophoto}{Sina Shaham} is a reseatch scientist at Meta. He
received the Ph.D. degree at the University of Southern California. He has also worked as a Data Scientist and a Software Engineer. His research interests include privacy and machine learning.
\end{IEEEbiographynophoto}

\begin{IEEEbiographynophoto}{Dinh C. Nguyen} is an assistant professor at the Department
of Electrical and Computer Engineering, The University of Alabama in
Huntsville, USA. He worked as a postdoctoral research associate at Purdue
University, USA from 2022 to 2023. He obtained the Ph.D. degree in computer
science from Deakin University, Australia in 2021. His research interests
include federated learning, Internet of Things, wireless networking, and
security. He has published over 70 papers on top-tier IEEE/ACM conferences
and journals such as IEEE JSAC, IEEE COMST, IEEE TMC, and IEEE IoTJ.
He is an Associate Editor of IEEE Transactions on Network Science and Engineering. He received
the Best Editor Award from IEEE Open Journal of Communications Society
in 2023.
\end{IEEEbiographynophoto}

\end{document}